%% file: main.tex
\documentclass[twoside]{article}
\input{./math_commands.tex}

\usepackage[accepted]{aistats2023}

\usepackage{graphicx}
\usepackage{hyperref}
\usepackage{url}
\usepackage{multirow}
\usepackage{caption}
\usepackage{subcaption}
\usepackage{xcolor}

\usepackage[round]{natbib}

\begin{document}

%

%

\twocolumn[

\aistatstitle{Probing Graph Representations}

\aistatsauthor{ Mohammad Sadegh Akhondzadeh \And Vijay Lingam \And  Aleksandar Bojchevski }

\aistatsaddress{  CISPA Helmholtz Center for Information Security } ]

\begin{abstract}
  Today we have a good theoretical understanding of the representational power of Graph Neural Networks (GNNs). For example, their limitations have been characterized in relation to a hierarchy of Weisfeiler-Lehman (WL) isomorphism tests. However, we do not know what is encoded in the learned representations. This is our main question. We answer it using a probing framework to quantify the amount of meaningful information captured in graph representations. Our findings on molecular datasets show the potential of probing for understanding the inductive biases of graph-based models. We compare different families of models and show that transformer-based models capture more chemically relevant information compared to models based on message passing. We also study the effect of different design choices such as skip connections and virtual nodes. We advocate for probing as a useful diagnostic tool for evaluating graph-based models.
\end{abstract}

\section{INTRODUCTION}
In this paper, we ask a deceptively simple question: \emph{What is encoded in the representations learned by graph-based models?}
For example, if we use a Graph Neural Network (GNN) to predict the toxicity of a molecule, will the hidden representation contain information about the number of hydrogen atoms or the presence of aromatic rings?
Prior works put constraints on the set of possible answers. For a GNN that is only as expressive as a 1-Weisfeiler-Leman test (1-WL), the information necessary to distinguish between graphs higher in the WL-hierarchy is provably not encoded. This fact has inspired a series of works that aim to build even more powerful and more expressive GNNs \citep{xu2018powerful,RFHLRG19Neural}. Nonetheless, just because a GNN is theoretically able to distinguish between two graphs, there is no guarantee that it will learn to do so in practice.
Further limitations such as over-smoothing \citep{OonoS20Graph} and over-squashing \citep{Alon21Bottleneck} stemming from the message passing paradigm impose additional constraints.
Graph Transformer models \citep{ying2021transformers, Kim2022PureTA} aim to alleviate these limitations using a variety of structural and positional biases with impressive results on downstream tasks. 
It is natural to ask if their success is partly due to their ability to better capture the relevant (e.g. chemical) knowledge.\looseness=-1

A trivial answer to our opening question is that the architecture, the dataset, the learning task, the optimization algorithm, and all other relevant design choices (e.g. data augmentation) jointly determine what the model learns to encode in the latent representations.
Since theoretically characterizing the effect of these choices is challenging, we tackle the question with a rigorous empirical analysis. Specifically, we adopt the \emph{probing} framework \citep{Belinkov2022ProbingCP}, which has proven useful in answering similar questions in the natural language processing domain (e.g. is the length of a sentence encoded in its representation).
The idea behind probing is simple. If we can reliably extract a given property (e.g. the presence of aromatic rings) from a given representation (e.g. the output of the penultimate layer in a GNN), we can conclude that the information about that property is encoded in the representation. Defining what it means to reliably extract a property leads to different flavors of probing. In its simplest form a so called linear probe evaluates whether a linear classifier can accurately predict the property given the fixed representations as input.

In this paper, we use this powerful framework to compare different families of models, e.g. traditional GNNs vs. Graph Transformers, and the effect of different modelling choices, e.g. using skip connections or virtual nodes. 
We focus on probing molecular representations since molecular prediction tasks are one of the main testbeds to benchmark graph-based models. We probe for chemical knowledge such as information about the atoms, the presence of important functional groups, properties related to the molecule's 3D structure, and other high-level properties.

To obtain a more complete picture we rely on three complementary probing strategies. 
We start with linear probing since the findings are the easiest to interpret.
Then, we employ an information-theoretic probing framework, which aims to estimate the Bayesian Mutual Information (BMI) between a representation and a property \citep{pimentel2021bayesian}. BMI generalizes classical MI and quantifies the amount of information a rational agent could extract given partial knowledge.
Finally, we use a paired probing strategy, where we intervene on the input, directly changing a property of interest (e.g. by deleting a functional group) to isolate its effect on the representation. In addition, we study the correlation between the richness of the representations space -- quantified via probing -- and the transferability of the representations to other downstream tasks.\looseness=-1

Unsurprisingly, models that perform better tend to encode more relevant information in their representations. Nonetheless, it is helpful to quantify the effect of different design choices, e.g. we show that models with skip connections capture more information.    
Interestingly, we find that even randomly initialized models can extract useful features, e.g. they encode as much information about certain properties as pretrained models.
This raises questions regarding the adequacy of common evaluation protocols.

We advocate for the use of probing as a diagnostic tool to better understand what is learned by our models and how they learn. For example, if certain properties are reliably extractable, even for out-of-distribution graphs, our model might be able to generalize better. It is especially important to verify whether the model learns properties that are known to be important based on domain knowledge.
We also investigate the acquisition of knowledge during training to understand which properties are more easily learned and how they relate to the downstream task. This highlights the potential of probing as a debugging tool to aid the development of better models. Similarly, probing can help us with the selection of (pretrained) models.
Finally, even though it is not the focus of this work, probing can surface properties that might be necessary to solve a certain task. If all accurate models inevitably learn a certain property it may be causally related to the prediction.

\section{PROBING METHODOLOGY}
Let $f: \vx \mapsto y$ be a model that maps an input $\vx$ to an output $y$. We consider functions $f$ that generate intermediate (hidden) representations of $\vx$. Let $f_l(\vx) = \vz, \vz \in \sR^d$ refer to the $d$-dimensional output of layer $l$ in a neural network such as a GNN or a Graph Transformer.
To simplify the exposition, we focus on the case where the input $\vx$ is a molecule, but our approach is easily applicable to other tasks. Similarly, $\vz$ may also refer to learned attention weights or any other intermediate output.

Our goal is to understand what kind of information is encoded in the representation $\vz$. To do so we use a probing dataset $\gD=\{f(\vx_i), p_i\}_{i=1}^N = \{\vz_i, p_i\}_{i=1}^N$ where $p_i$ is a property of interest. For example, $\vx_i$ is a molecule, $\vz_i$ is the output of the second layer of a GNN, and $p_i$ is the presence of aromatic rings.
The probing dataset is independent of the datasets used to train and evaluate the original model $f$. 
%
We can efficiently compute many relevant properties, which means that given any unlabeled set of molecules we can automatically create a probing dataset $\gD$. For other properties we will rely on existing annotated (labeled) datasets.

\subsection{Probing Properties}
\label{sec:properties}
We focus on five different types of properties when probing molecular representations with the goal of evaluating whether the model learns chemically relevant information.

\textbf{Atom Counting.} 
For the first set of properties, we simply consider the number of different atoms in a molecule. We compute the number of Carbon, Nitrogen, and Oxygen atoms since they are the most common atoms in organic compounds. Therefore, here $p_i \in \sN, p_i \geq 0$.

\textbf{Meaningful Substructures.}
Second, we rely on chemical domain knowledge to find meaningful substructures in a molecule. We explore eight different substructures including: Aromatic Carbocycles, Aromatic Rings, Saturated Rings, Aniline, Benzene, Bicycle, Ketone, Methoxy, Para Hydroxylation, and Pyridine. These substructures are often referred to as functional groups since they usually have their own characteristic properties, regardless of the other atoms present in a molecule. For example, as the name suggests, the presence of aromatic rings changes the odor of a molecule. Using the RDKit tool \citep{Landrum2016RDKit} we compute
whether a molecule contains one of these functional groups, thus $p_i \in \{0, 1\}$. 
To illustrate the principle, we only probe for these eight properties, but our approach scales to many more properties. Using RDKit, we can compute over 80 different functional groups. Similarly, RDKit can compute other chemically relevant properties such as the number of radical electrons or the number of rotatable bonds.

\textbf{Molecular Properties.}
Next, we probe the model representations for high-level molecular properties that are different from the property/label used to train the original model $f$. For example, using the PCQM4Mv2 dataset \citep{hu2020ogb}, models are trained to predict the HOMO-LUMO energy gap of molecules (calculated with density functional theory) given their 2D molecular graphs.
For probing we first consider the MoleculeNet benchmark \citep{Wu2017MoleculeNetAB} which contains multiple datasets with various tasks derived from different properties such as blood-brain permeability, toxicity, and lipophilicity. Some of the properties are continuous, $p_i \in \sR$, others are binary or categorical $p_i \in \{0, \dots, C\}$ where $C$ is the number of categories. 
%
In contrast to the previous types (atom counting and meaningful substructures), which we can compute for any molecule, these  properties require labeled datasets. Therefore, this setup is closely related to transfer learning. Our probes can be seen as evaluating whether the representations trained on a source task transfer to a different (target) task.\looseness=-1



\textbf{3D Properties.}
Since 3D geometric information plays an important role in determining the function of a molecule \citep{GilmerSRVD17MessagePassing}, we explore whether the learned representations capture any properties related to the 3D-structure of the molecules.
Note, the models are trained using only the molecular graph as input without any 3D input. Using RDKit, we compute five 3D properties: Asphericity, Radius of Gyration, Sphericity Index, NPR1 (normalized first principal moments ratio), and PMI1 (smallest principal moment of inertia). All of these are continuous ($p_i \in \sR$).

\textbf{Odor Properties.}
Finally, we probe the the learned representations for information about the odor (smell) of a molecule, e.g. sweet or woody (32 smells in total), using the Pyrfume public data archive\footnote{Each dataset and the corresponding reference can be found on \url{https://github.com/pyrfume/pyrfume-data}.}. Here $p_i \in \{0, 1\}$.

\subsection{Probing Strategies}
\label{sec:probing_strategies}
\textbf{Linear Probing.}
The goal of probing is to evaluate the "extractability" or "readability" of a property from the representation. The standard approach is to train a separate model, called a probe, to \emph{predict} the property $p_i$  given the fixed representations  $\vz_i$. Specifically, the probing dataset $\gD$ is split into a train and test set, the probe is trained on the train set, and its performance is evaluated on the test set. Good test performance is taken as evidence that the representation contains information about the property. Low performance indicates that the property is either not present in the representations or not usable. The idea of usability is prominent in the literature.
The advocates of linear probing \citep{AlainB17Understanding} argue that the probe model should be simple, e.g. a logistic regression (or linear regression for continuous properties),  since this means that the information can be easily extracted and used in subsequent processing (e.g. in subsequent layers).
The advocates for more complex probes \cite{PimentelVMZWC20Information} argue they are better since the information about the property may be non-linearly encoded in the representation. Yet, the good performance of
non-linear probes may come from overfitting (memorization of spurious correlations). To overcome these limitations, various control tasks have been proposed: comparing the performance to a majority baseline, random representations, randomization of the properties, or the use of minimum description length. 
Despite its limitations, in this paper we use linear probing since the results are more interpretable. If a linear probe has good performance then there exists a hyperplane in the representation space that separates the inputs based on their properties.  



\textbf{Bayesian Probing.}
Since the goal is to measure information about a property, it is natural to resort to information-theoretic concepts such as Mutual Information (MI). For two random variables $Z$ and $P$, the mutual information is defined as $I(Z;P)=H(Z) - H(Z\mid P)$ where $H$ is the (conditional) entropy.
For example, \citet{PimentelVMZWC20Information} argue that training probe models can be seen as estimating the MI between the representations and the property.
However, recalling the data processing inequality we realize that there is a conceptual problem with using MI. For a set of random variables that form a Markov chain $(X \xrightarrow[]{} Z \xrightarrow[]{} P)$ we have $ I(X;P)\geq I(Z;P)$. That means that any subsequent processing of the input $\vx$, i.e. applying any function including $f_l(\vx)$, can only reduce information. To tackle this issue \citet{XuZSSE20Usable} propose a theory of usable information under computational constraints called $\gV$-information. \citet{pimentel2021bayesian} generalize this idea to the so called Bayesian Mutual Information (BMI). 
Both of these notions generalize the classical Shannon entropy, and by extension the Shannon MI, breaking the data processing inequality. 
In this paper, we employ BMI since it fixes some of the issues with $\gV$-information (see \citet{pimentel2021bayesian} for a detailed discussion).

BMI estimates the information a rational agent could obtain from a random variable given only \emph{partial} knowledge of its distribution. 
Since discussing the theory of BMI is beyond the scope of this paper, we only recall some of its relevant properties.
First, it is symmetric, $I(Z;P)=I(P;Z)=H(P)-H(P|Z)$. 
Second, it depends on the (size of the) probing dataset used to train the agent. Therefore, we perform incremental probing where we gradually increase the size. 
Similar to \citet{pimentel2021bayesian} we specify the prior beliefs of the agent as a Categorical distribution (which has a Dirichlet prior on top), and we approximate the agent's posterior beliefs with the maximum-a-posteriori estimate (assuming a Gaussian prior on the weights). Operationally, to estimate BMI we compute $H(P)$ in closed-form (see \citet{pimentel2021bayesian}), and  we train linear predictors on increasingly larger probing datasets using the predicted probabilities to compute the conditional entropy $H(P\mid Z)$.

\begin{figure}[t]
    \centering
    \includegraphics[width=0.4\textwidth]{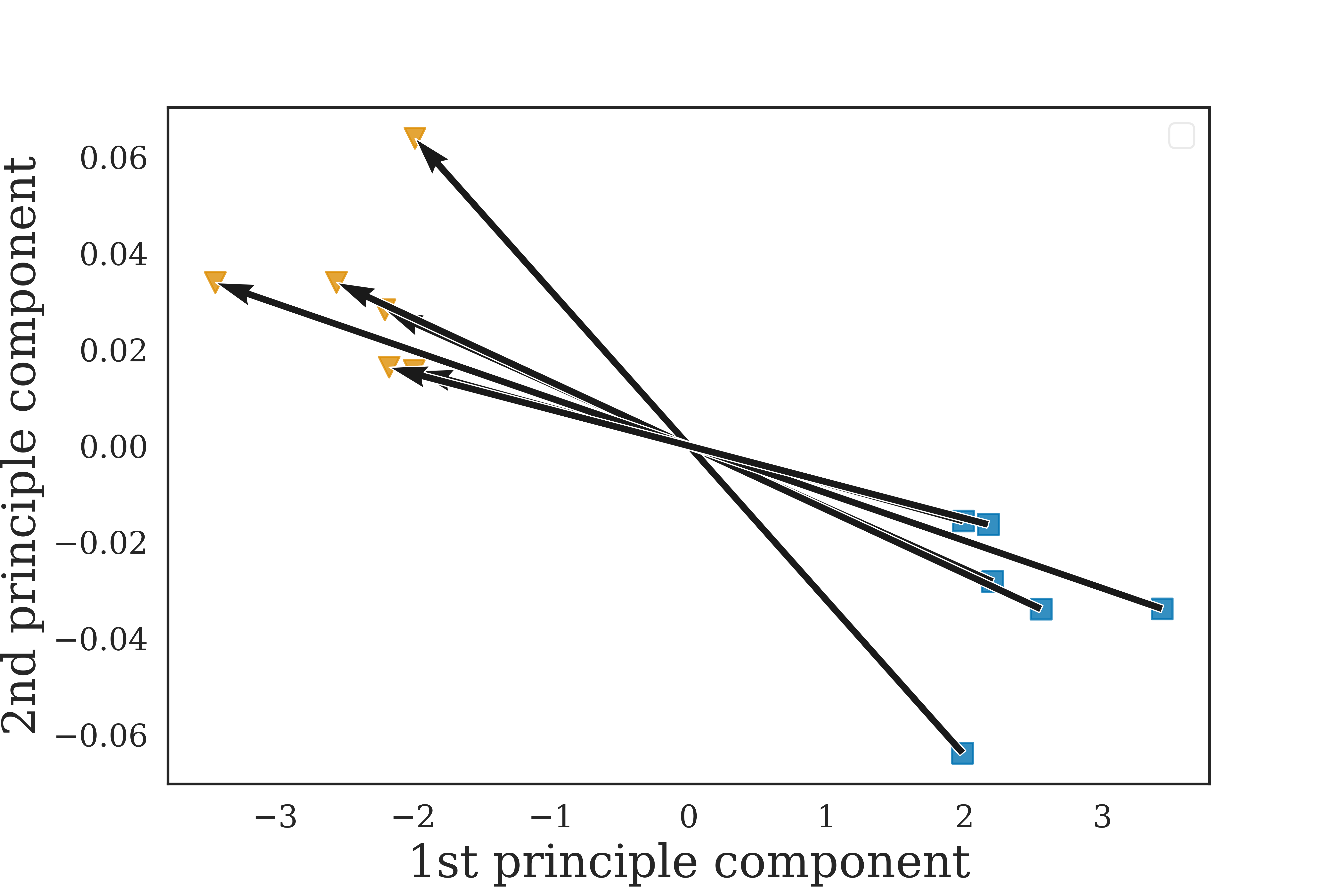}
    \includegraphics[width=0.4\textwidth]{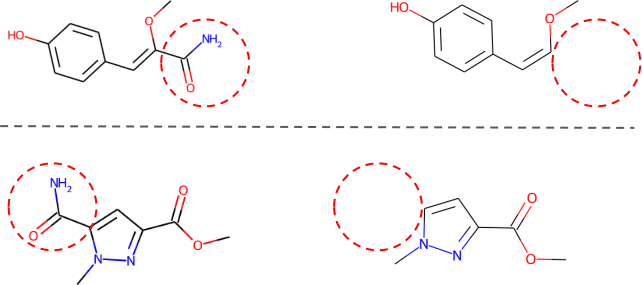}
    \caption{Pairwise probing when removing the amide functional group (dotted circle). Arrows connect each source molecule to its target. The directions are highly aligned. The first principal component is sufficient to separate molecules with ({squares}) and without ({triangles}) an amide. }
    \label{fig:pairwise_illustration}
\end{figure}

\textbf{Pairwise Probing.} To better isolate the effect of a certain property on the representation we propose pairwise probing. The main idea is to create a probing dataset $\gD=\{(\vx_i, \vx'_i)\}_{i=1}^N$ of pairs of molecules such that each pair is as similar as possible, but differing \emph{only} in the property of interest. We demonstrate this idea for properties corresponding to meaningful substructures (functional groups).
Assume that the property $p \in \{0, 1\}$ is binary. First, we create a set $\gS$ of molecules such that $p_i=1$ for each $\vx_i \in \gS$. In other words, all of the molecules in the set $S$ contain a given substructure (e.g. all molecules contain a Benzene ring). Then, for each $\vx_i$ we create its corresponding pair molecule $\vx'_i$ by removing the substructure of interest.
We make sure that the resulting molecule is still valid and add hydrogen atoms to fill the capacity of the atom(s) previously connected to the functional group.
This means that by design for each $\vx'_i$ the property is $p'_i=0$, and that each pair is $(\vx_i, \vx'_i)$ is minimally different except for the property of interest. We call $\vx_i$ the source molecule and $\vx'_i$ the target.
\autoref{fig:pairwise_illustration} shows an example of this process for the amide functional group. 

We can use the pairwise property in several ways. 
First, inspired by the semantic analogy for word embeddings \citep{DBLP:conf/nips/MikolovSCCD13}, we evaluate whether the vectors connecting the representations of the source molecules to their targets, i.e. $\vv_i = f(\vx_i) - f(\vx')_i = \vz_i - \vz'_i$ are aligned with each other. To do so, we compute the cosine similarity between each $(\vv_i, \vv_j)$ pair. If the similarity is high on average it means that there is a consistent direction in the representation space that identifies the property of interest. Similar to linear probing, the hyperplane perpendicular to this direction can separate source from target molecules.

Next,  we run a principal component analysis (PCA) to identify whether the directions of highest variation are aligned with the property. We center the representations with $\hat{\vz}_i=\vz_i - \vm_i$, and $\hat{\vz}'_i=\vz'_i - \vm_i$, where $\vm_i = 0.5 \cdot (\vz_i+\vz'_i)$. This has the effect of moving the representation of each molecule such that all difference vectors $\vv_i$ cross the origin at the midpoint. This centering helps to better isolate the main source of variation. Then we run standard PCA on the matrix $\mZ = [\hat{\vz}_i; \hat{\vz}'_i] \in \sR^{2N \times d}$ containing all centered representations. This approach is illustrated on \autoref{fig:pairwise_illustration}. We visualize the projection of the representations onto the top two principal components, isolating the relevant subspace for the property. 
We see that even a single principal component is sufficient to separate the molecules that contain the \textit{amide} functional group from those that do not.\looseness=-1

Finally, another way to use the paired dataset is to discard the pairing information and simply use $\gD$ with the previous two probing strategies (linear probing and Bayesian MI). The benefit is that, unlike before, the dataset is now balanced w.r.t. $p_i$, making the probe models easier to train.

\begin{figure}[t]
    \centering
    \includegraphics[width=0.48\textwidth]{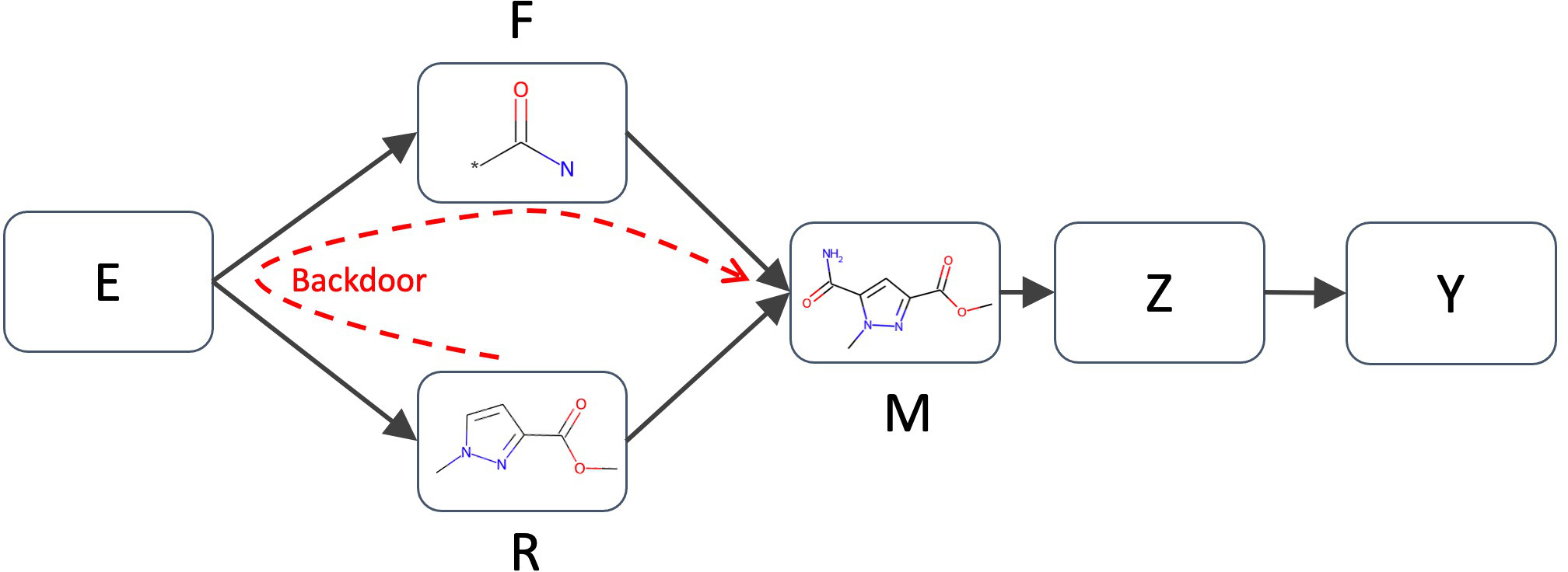}
    \caption{
    Causal DAG showing the dependencies between the environment (E), the functional group (F), the remaining structure (R), the molecule (M), the representation (Z), and the predicted label (Y) of the molecule.
    }
    \label{fig: causal_graph}
\end{figure}

As suggested by \citet{Amini2022NaturalisticCP}, our paired probing strategy can be interpreted as causal probing by considering each $\vx'_i$ as the counterfactual of $\vx_i$ after intervention on the property. If we make assumptions for the Structural Causal Model (SCM) we can estimate the causal effect of the property. Similar to \citet{yanglearning}, we assume our generative process is based on the causal DAG represented on \autoref{fig: causal_graph},
where $E, F, M, Z, R$ and $Y$ are random variables corresponding to the environment, the functional group, the molecule, the representation, the remaining structure, and the high-level property of the molecule (label)  respectively. 
Using observational data to compute  $p(Z \mid F = 1)$ we obtain biased estimates since there is a backdoor path $F \leftarrow E \rightarrow R \rightarrow M \rightarrow Z$ from $F$ to $Z$. However, by intervening on $P$ as we propose we can block the backdoor path. Then, using the paired dataset we can estimate the average treatment effect $p(Z \mid \text{do} (F=1)) - p(Z \mid \text{do} (F=0))$ computing $\vv_\text{ate}=\frac{1}{N} \sum_{i=1}^N \vv_i=\frac{1}{N}\sum_{i=1}^N \vz_i - \frac{1}{N}\sum_{i=1}^N\vz'_i$. 

To be clear, the SCM relies on strong assumptions and we do not argue that it necessarily represents the true data-generating distribution. Nonetheless, the implications given those assumptions are insightful -- the average treatment effect $\vv_\text{ate}$ captures the direction from molecules with a given property to molecules without it. 
We verify whether $\vv_\text{ate}$ is aligned with the source-target directions by computing $C_\text{ate}^\text{pair}$ the average cosine similarity between $\vv_\text{ate}$ and $\vv_i$.
Using pairwise probing we can also study the (causal) relationship between functional groups and other high-level properties. In~\autoref{sec:appendix_causal_smell} we study the effect on odor. Our findings match well-known results from chemistry.

\input{./figures/table_linear}
\input{./figures/table_3d_counting}
\section{EXPERIMENTS}
\label{sec:probin_results}
We study popular message-passing GNNs and Graph Transformer architectures.\footnote{Our code is publicly available at \url{https://github.com/msadegh97/probing-graph-representation}.} For the message-passing models, we use GCN \citep{kipf2016semi} and GIN \citep{xu2018powerful}. GIN is theoretically more expressive.
For the transformer models, we study Graphormer \citep{ying2021transformers}, GRPE \citep{park2022grpe}, GraphGPS \citep{rampavsek2022recipe}, and TokenGT \citep{Kim2022PureTA}. We train all models on the large scale PCQM4Mv2 dataset \citep{hu2021ogblsc}, and we probe their hidden representations for various properties. As a baseline, we also probe the raw features (using sum of input features across nodes as a representation), and Morgan fingerprints \citep{morgan1965generation} -- manually crafted representations based on domain knowledge. 
We create the probing dataset $\gD$ from the PCQM4Mv2 dataset. We randomly sample 100K molecules from the training and 20K molecules from the validation set, thus maintaining a scaffold-split. For binary properties we report the ROC AUC score and for ordinal/continuous properties we report the R2 score. 

\textbf{Hyperparameters and  Reproducibility.}
To pretrain all models used in our experiments, we rely on the authors' code and the checkpoints released by them. For models without checkpoints, we use the sweep over the hyperparameters ranges suggested by the authors. Our probes utilize linear regression models. See \autoref{sec:appendix_train_eval_linear_probes} for more details.

%

%
%

\begin{table}[!b]
\vspace*{-10pt}\caption{Linear probing for high-level molecular properties.}
\label{tab:transfer}
\resizebox{0.48\textwidth}{!}{
\begin{tabular}{ccccccccc|c}
\hline
\textbf{Dataset}         & \textbf{BBBP}  & \textbf{ToxCast} & \textbf{Sider} & \textbf{ClinTox} & \textbf{BACE}   & \textbf{AVG}   \\ \hline
\textbf{Molecules}       & 2,039 & 8,575   & 1,427 & 1,478  & 1,5113 &       \\
\textbf{Tasks}           & 1     & 617     & 27    & 2      & 1      &       \\ \hline
\textbf{Raw} & 66.8 & 60.5 & 56.2 & 60.6 & 75.1 & 66.3  \\
\textbf{Morgan} & 63.2 & 54.1 & 60.5 & 57.4 &  75.0 & 63.4 \\ \hline
\textbf{GCN}     & 56.5  & 57.0    & 55.2  & 64.9   & 71.1   & 63.0  \\
\textbf{GIN}     & 58.8  & 55.7    & 53.2  & 57.0   & 77.1   & 62.1  \\ \hline
\textbf{Graphormer}      & 59.6  & 59.2    & 57.5  & 78.5   & 63.0   & 67.2  \\
\textbf{GRPE}            & 64.9  & 61.1    & 55.7  & 82.8   & 77.9   & 70.4  \\
\textbf{GraphGPS}        & 63.9  & 60.0    & 60.7  & 72.7   & 74.5   & 68.5  \\
\textbf{TokenGT} & 57.1	& 59.3	  & 58.4  &	88.5   & 75.2   & \textbf{70.6}  \\ \hline
\end{tabular}
}
\end{table}

\textbf{Impact of the Architecture.}
In \autoref{tab:liinear_probing_models} we observe that on average transformer-based models tend to capture more information about the functional group properties -- they learn more chemically relevant knowledge. Even though GIN is theoretically more expressive than GCN it performs significantly worse. The raw features perform surprisingly well. One reason is that they explicitly encode some of the relevant information, e.g. whether an atom is part of a ring. 
In \autoref{tab:atom_counting} we report the R2 score for atom counting and 3D structure properties.
Values closer to 1 are better and N/A indicates that the probe was not able to converge.
Again, we see that the transformer-based models perform significantly better. Interestingly, some models are able to capture a good amount of 3D information even though no explicit 3D structure is provided during training.

We continue with linear probing for high-level molecular properties using the MoleculeNet benchmark.
In \autoref{tab:transfer} we show a subset of the properties and the average performance on all $8$ properties (see~\autoref{sec:app_additional_probing_perf} for details).
Interestingly, the raw features outperform both the Morgan fingerprints and the GNN models. Transformer-based models are still significantly better.

Finally, we present the results for a subset (5/32) of the the odor properties and the overall average in \autoref{tab:app_smell_main}. We provide the detailed explanation and  the full table in \autoref{sec:app_smells}. 
We see that, again, the representations learned by transformer-based models encode more information about odors on average. GCN outperforms GIN even though it is less expressive. Here, raw features and Morgan fingerprints are significantly worse.
Certain odors (e.g. see sulfurous in \autoref{tab:app_smell}) are reliably encoded by all models.

\begin{figure}[!b]
\centering
\includegraphics[width=0.48\textwidth]{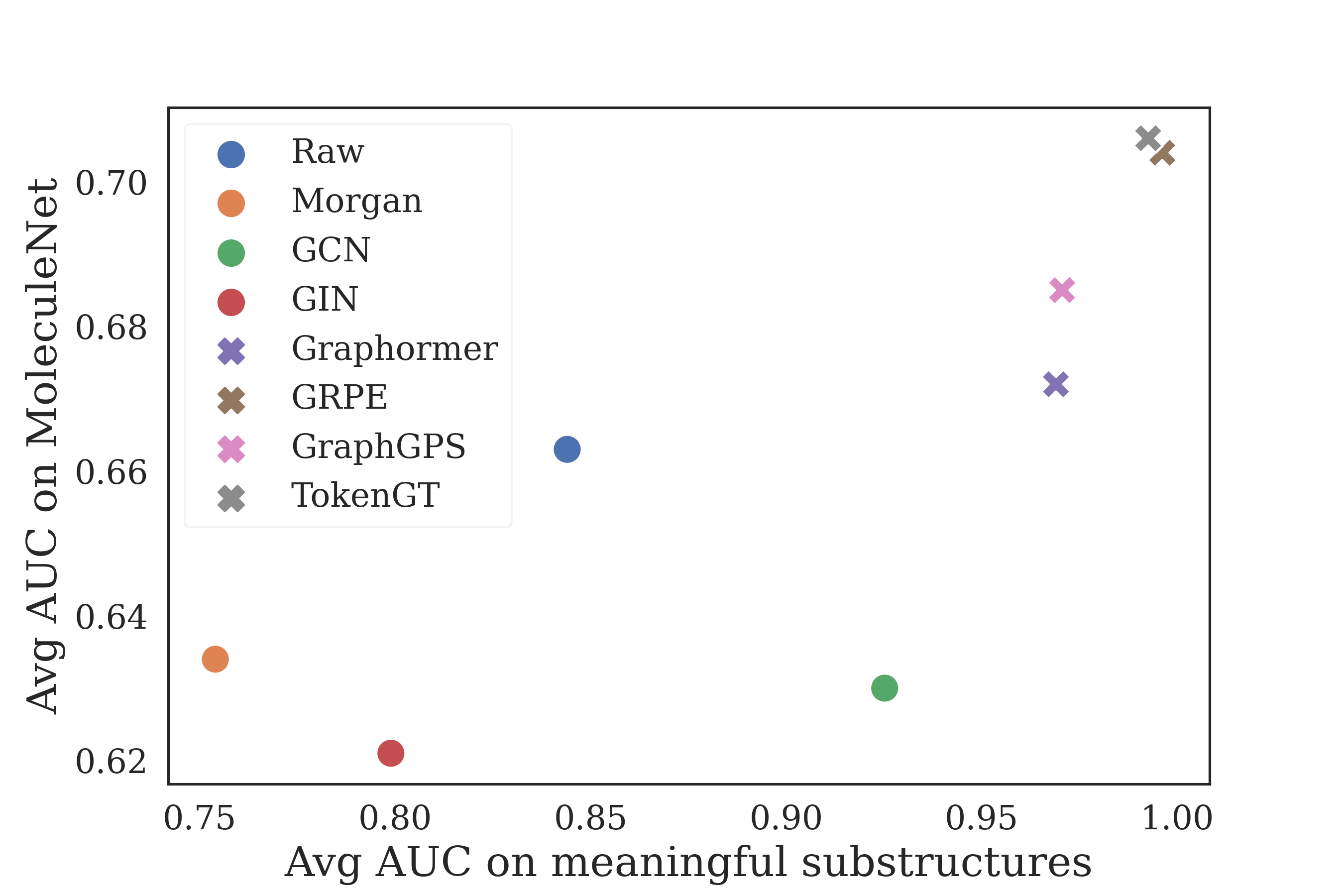}
\caption{Substructures vs. high-level properties.}
\label{fig:correlation}
\end{figure}

Overall, transformer-based models perform better on average. This bolsters our hypothesis that they learn richer representations allowing them to better transfer to other tasks. Providing further evidence for this, in \autoref{fig:correlation} we see a strong correlation between the average probing performance w.r.t. the functional groups and the average performance on MoleculeNet.

\begin{table}[!t]
\centering
\caption{Linear probing performance (AUC) for odors.}
\label{tab:app_smell_main}
\resizebox{0.48\textwidth}{!}{
\begin{tabular}{c|ccccc|c}
\hline
\textbf{Model}      & \textbf{Apple} & \textbf{Balsamic} & \textbf{Burnt} & \textbf{Caramellic} & \textbf{Cheesy} & \multicolumn{1}{l}{\textbf{AVG}} \\ \hline
\textbf{Raw}        & 0.52           & 0.57              & 0.55           & 0.55                & 0.60             & 0.59                                 \\
\textbf{Morgan}     & 0.55           & 0.64              & 0.50           & 0.52                & 0.52                  & 0.57                                 \\ \hline
\textbf{GCN}        & 0.91           & 0.88              & 0.90           & 0.93                & 0.92                   & 0.85                                 \\
\textbf{GIN}        & 0.80           & 0.88              & 0.85           & 0.77                & 0.79                   & 0.79                                 \\
\textbf{Graphormer} & 0.88           & 0.87              & 0.85           & 0.84                & 0.90                     & 0.83                                 \\
\textbf{GRPE}       & 0.88           & 0.91              & 0.91           & 0.88                & 0.88                   & 0.85                                 \\
\textbf{GraphGPS}   & 0.89           & 0.87              & 0.86           & 0.92                & 0.89                  & \textbf{0.86}                        \\
\textbf{TokenGT}    & 0.90           & 0.94              & 0.90           & 0.88                & 0.87          & 0.85    \\ \hline
\end{tabular}
}
\end{table}

\begin{figure}[!b]
\centering
\includegraphics[width=0.48\textwidth]{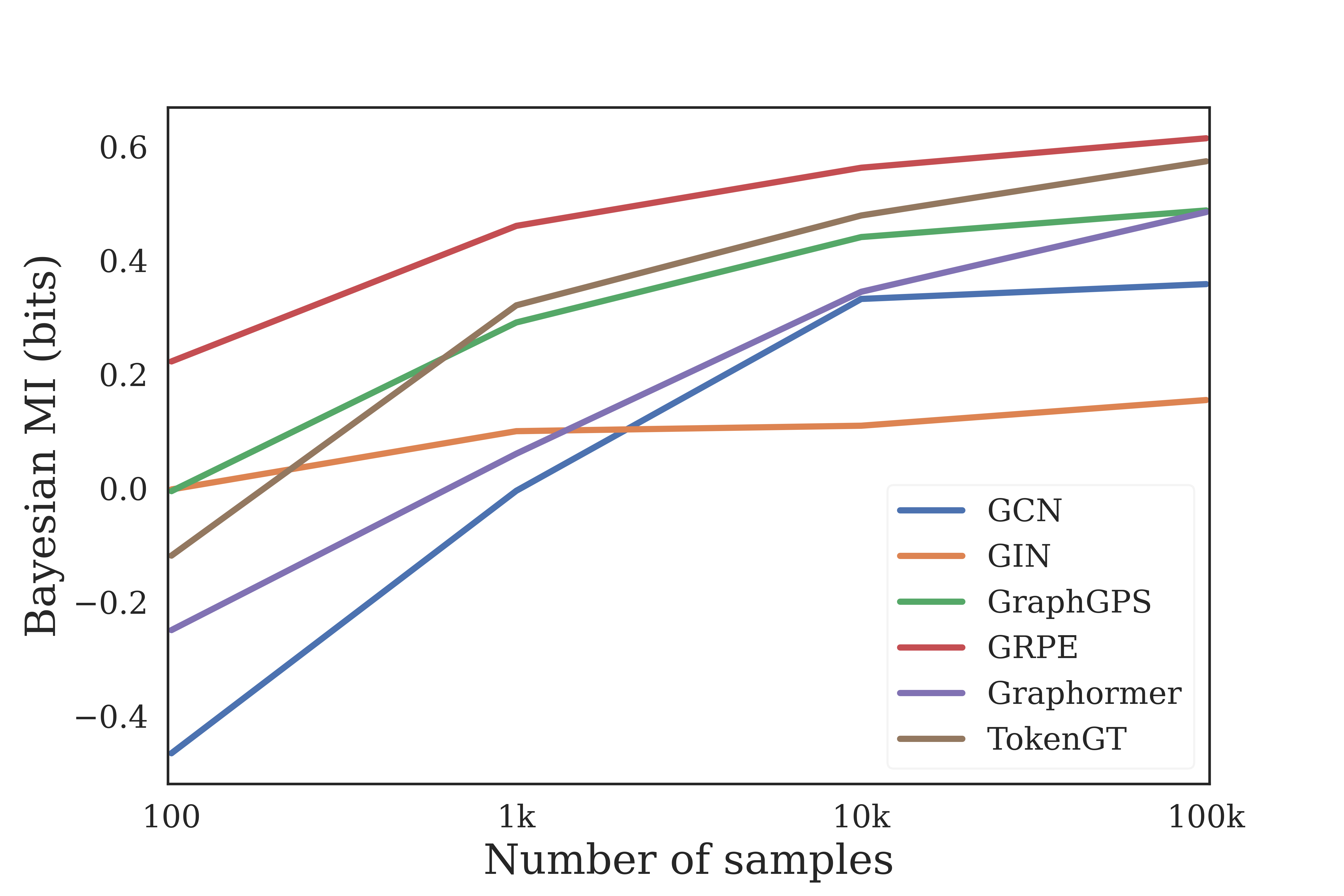}
\caption{Probing with Bayesian mutual information. }
\label{fig:bayesian_probing}
\end{figure}


\begin{figure*}[!t]
    \centering
    \includegraphics[width=\textwidth]{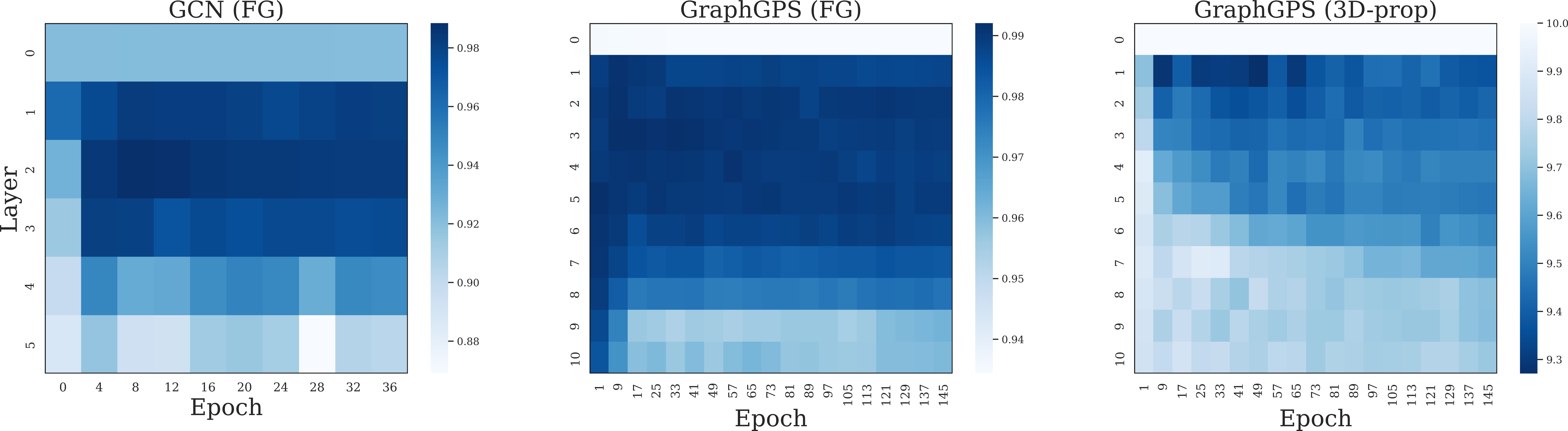}
    \captionof{figure}{Average probing performance on functional groups and 3D properties across epochs and layers.}
    \label{fig:training_heatmap}
\end{figure*}

\begin{table}[!t]
\captionsetup{type=table}
\captionof{table}{Random vs. pretrained models.\label{tab:random}}
    \centering
    \resizebox{0.45\textwidth}{!}{
    \begin{tabular}{c|cc|cc}
                             & \multicolumn{2}{c|}{\textbf{FG (AUC)}} & \multicolumn{2}{c}{\textbf{3D-Prop (R2)}} \\\hline
    \textbf{Model}           & \textbf{Random} & \textbf{Train} & \textbf{Random}   & \textbf{Train}   \\\hline
    \textbf{GCN}           & 0.905                & 0.925              & N/A                    & N/A                  \\
    \textbf{GIN}           & 0.897                & 0.799              & -0.552                 & -139.4               \\ \hline
    \textbf{Graphormer}      & 0.975                & 0.965              & -0.279                 & 0.417                \\
    \textbf{GRPE}            & 0.996                & 0.995              & 0.186                  & 0.610                \\
    \textbf{GraphGPS}        & 0.981                & 0.971              & 0.541                  & 0.505                \\
    \textbf{TokenGT} & 0.963                & 0.993              & 0.083                  & 0.589               
    \end{tabular}
    }
\end{table}


\textbf{Bayesian Mutual Information.}
Next, we use the Bayesian Mutual Information (BMI) probing strategy. We plot the results in \autoref{fig:bayesian_probing}.
Increasing the size of the probing dataset leads to an expected increase in information as measured by BMI.
Interestingly, GIN performs surprisingly well for small probing datasets (100 and 1000 samples), however, its performance plateaus as the Bayesian agent gets access to more data. This is unlike the other models which can take advantage of the increased size to better identify the property of interest. The ranking of models is stable, with transformer-based model being better than GNNs.


\textbf{Randomly Initialized Models.}
We also compare the performance of representations obtained from randomly initialized models (without pretraining) and the models pretrained on PCQM4Mv2. In \autoref{tab:random} we see that the untrained models are surprisingly good feature extractors. For the functional group properties they even perform on par or better than pretrained models. The random GraphGPS model also perform well for 3D properties. These results highlight the usefulness of the models' inductive biases.

\textbf{Training Dynamics.} Next, we study how the amount of encoded information changes during training and across the representations learned in different layers. 
In \autoref{fig:training_heatmap} we plot a 2D heatmap for different epochs (horizontal) and layers (vertical). Each cell shows the average linear probing performance. Darker shade is better.
For both GCN and GraphGPS we see that already in the first layer we have significantly more information compared to the input features (layer 0). This tells us that accounting for the graph structure is beneficial.
The information is maintained in the intermediate layers, and is slowly removed as we move towards the final layer. One reason for this is that the representations in the final layer are becoming more specialized in order to solve the original task.

The functional group properties (left and middle column) are learned right away and the performance does not significantly change over epochs. For GraphGPS the performance is high even for epoch 0, i.e. for the randomly initialized model before any training is done, while for GCN there is a noticeable increase from epoch 0 to epoch 1.
For the 3D properties we see that in the first layer the performance starts high and slightly decreases over time. However, the performance in the subsequent layers is significantly lower at the start and it increases over time. Again, the layers close to the output encode less information.

\begin{table}[!b]
    \centering
    \caption{The effect of the pre-training dataset.}
    \label{tab:dataset_prob_mol}
\resizebox{0.48\textwidth}{!}{

\begin{tabular}{c|ccccc|c|c}
\hline
\textbf{Model} & \textbf{Dataset}    & \textbf{BBBP} & \textbf{ToxCast} & \textbf{Sider} & \textbf{ClinTox} & \textbf{BACE} & \textbf{AVG} \\ \hline
\multirow{3}{*}{\textbf{GCN}}            & \textbf{Molhiv}     & 62.5                           & 59.4                              & 60.1                            & 68.9                              & 79.6                           & \textbf{66.1}                          \\
                                & \textbf{Molpcba\_s} & 65.8                           & 61.0                              & 59.1                            & 79.0                              & 73.9                           & \textbf{67.8}                          \\
                                & \textbf{Molpcba}    & 65.0                           & 61.4                              & 59.4                            & 71.4                              & 71.7                           & 65.8                          \\ \hline
\multirow{3}{*}{\textbf{GIN}}            & \textbf{Molhiv}     & 56.4                           & 54.3                              & 55.2                            & 63.7                              & 62.5                           & 58.4                          \\
                                & \textbf{Molpcba\_s} & 62.0                           & 63.0                              & 60.3                            & 69.9                              & 71.1                           & 65.3                          \\
                                & \textbf{Molpcba}    & 62.1                           & 63.5                              & 60.0                            & 62.7                              & 74.5                           & 64.5                          \\ \hline
\multirow{3}{*}{\textbf{V-GCN}}          & \textbf{Molhiv}     & 59.5                           & 56.8                              & 58.5                            & 68.1                              & 74.4                           & 63.5                          \\
                                & \textbf{Molpcba\_s} & 64.7                           & 60.1                              & 58.4                            & 69.9                              & 72.2                           & 65.1                          \\
                                & \textbf{Molpcba}    & 62.5                           & 60.3                              & 60.2                            & 71.1                              & 69.6                           & 64.8                          \\ \hline
\multirow{3}{*}{\textbf{V-GIN}}          & \textbf{Molhiv}     & 55.4                           & 54.8                              & 56.0                            & 69.2                              & 66.7                           & 60.4                          \\
                                & \textbf{Molpcba\_s} & 62.8                           & 62.7                              & 58.8                            & 67.8                              & 77.4                           & 65.9                          \\
                                & \textbf{Molpcba}    & 64.7                           & 65.2                              & 61.4                            & 69.2                              & 78.0                           & \textbf{67.7}                          \\ \hline 
\end{tabular}

}
\end{table}

\textbf{Impact of the Pretraining Dataset.}
Next, we assess the impact of the pretraining dataset on the usability of the learned representations. 
For a fair comparison, we use the Molhiv and Molpcba datasets \citep{hu2020ogb} for pretraining since both contain small molecules with similar molecule statistics (see~\autoref{sec:appendix_molecule_stats}). We also select a random subset of Molpcba with the same size as Molhiv (called Molpcba\_s). This will help us disentangle the effect of the dataset size and the effect of multitask learning -- in Molhiv we predict a single label and in Molpcba we predict 128 labels.
We perform linear probing on the MoleculeNet benchmark and show the results in \autoref{tab:dataset_prob_mol}. We see that the performance when pretraining on Molpcba tends to be higher on average for most models. This is true even for the smaller subset and indicates that multitask learning might be beneficial since it results in a richer representation space.

\begin{figure}[!b]
        \centering
        \includegraphics[width=0.48\textwidth]{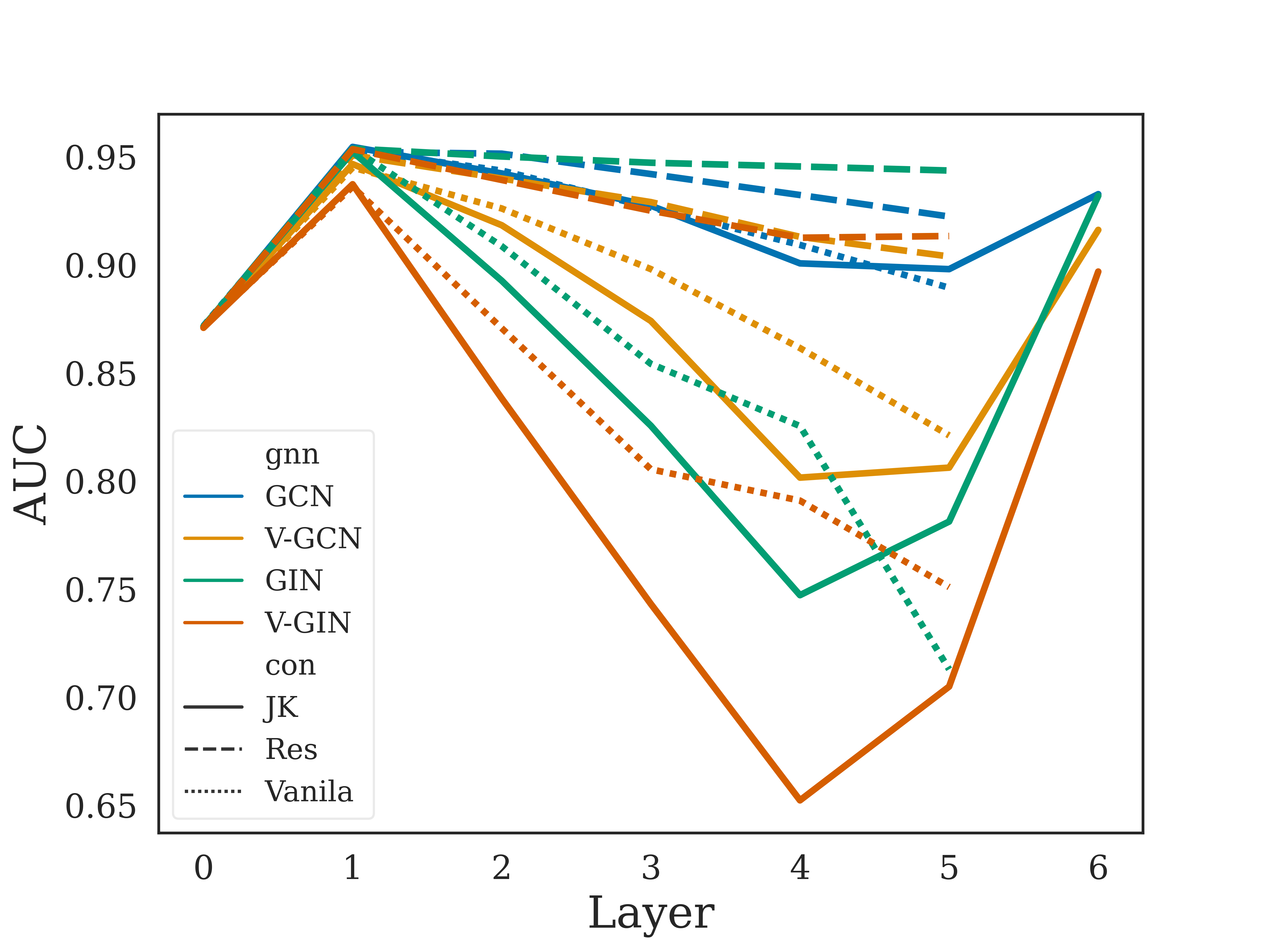}
        \caption{Impact of architectural design choices: jumping knowledge (JK) and residual connections (Res).}
        \label{fig: abl-DC}
\end{figure}

\begin{figure*}[!t]
    \centering
    \includegraphics[width=0.88\textwidth]{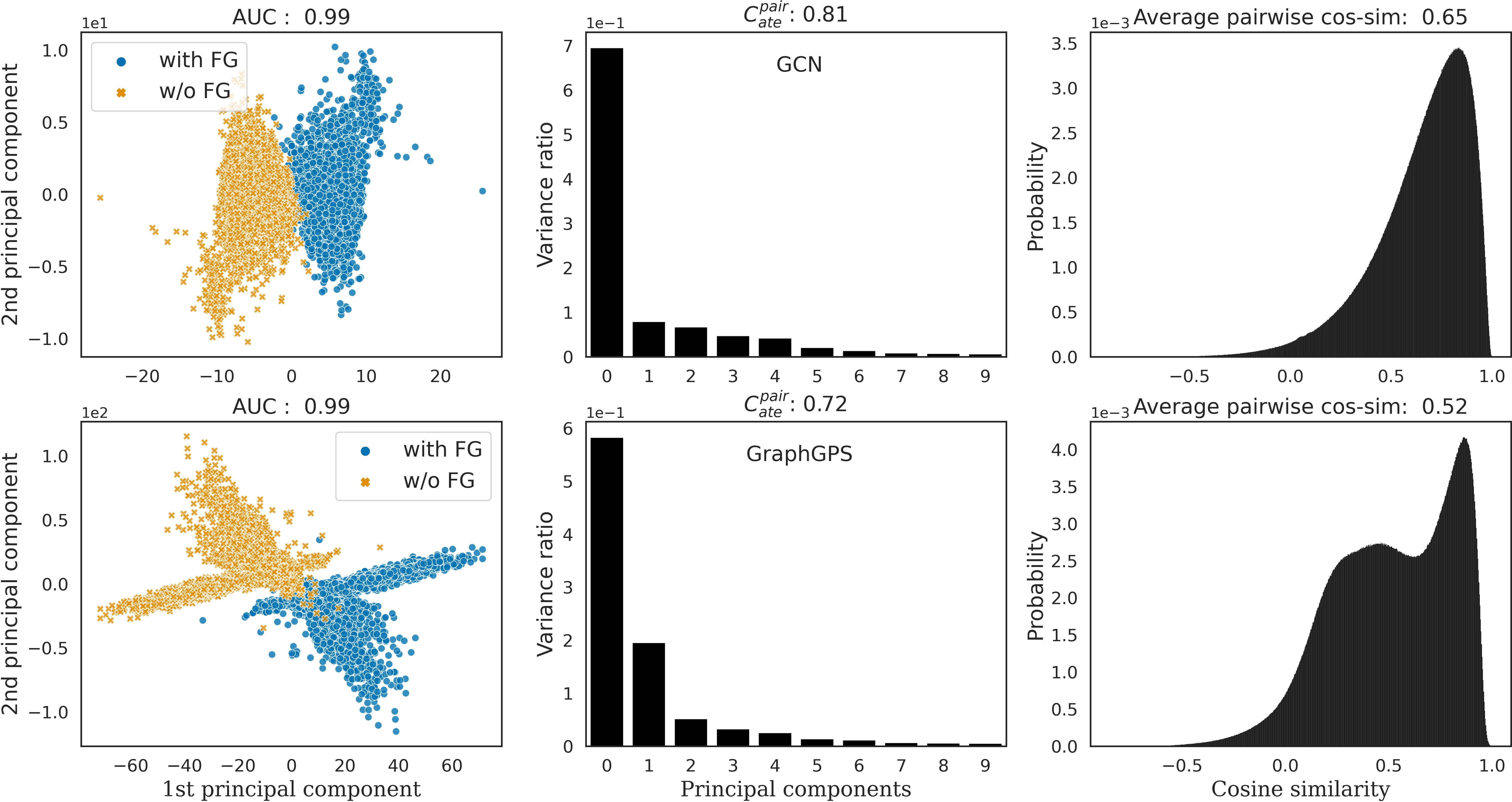}
    \caption{Pairwise probing for GCN (top) and GraphGPS (bottom). We show the projection on the top two principal components (left), the ratio of explained variance per component (middle), and the pairwise cosine similarity (right).}
    \label{fig:pca_plot}
\end{figure*}

\textbf{Impact of Design Choices.}
We study the impact of different design choices for the architecture of a GNN model. First, we consider residual connections (Res) and jumping knowledge (JK).
For this experiment we pretrained on the Molhiv dataset.
\autoref{fig: abl-DC} shows the average linear probing performance for all the meaningful substructure properties for networks trained with and without JK, and with and without Res. We consider the representations learned in each layer. We see that that using Res (dashed lines) helps preserve more information about the property across different layers. Compared to the vanilla model (dotted lines) where the information steadily drops with each layer (see also findings on training dynamics). The representations learned by the models using JK (solid lines) encode less information than the respective vanilla models, except at the last layer. This is expected since the last layer is a concatenation of all previous layers.
For all models and variants, the first layer (after one round of message passing) encodes the most information about a property.
We also study the impact of two pooling strategies, namely the default mean pooling (GCN, GIN) vs. virtual nodes (V-GCN, V-GIN). We see that compared to the representations of the virtual node using mean pooling results in representations that encode significantly more information about functional groups.\looseness=-1

\textbf{Pairwise Probing.}
Finally, we perform our pairwise probing strategy with the goal of explicitly isolating the effect of a certain property on the learned representations. Here we focus on the presence of meaningful substructures (functional groups).
We select one representative model from each family, namely GCN (top) and GraphGPS (bottom), and we show the results in \autoref{fig:pca_plot} for the nitro group. See \autoref{sec:appendix_pairwise_probing} for other models and properties.
In the left column we set the projection of the representations onto the top two principal components (PCs). We see that even a single principal component is sufficient to separate the source molecules (with nitro) from their counterparts (without nitro). This means that the direction of highest variance in the representation space is highly aligned with the property of interest.
The middle column in \autoref{fig:pca_plot} shows the ratio of the captured variance for each PC. We see that the first PC captures more than half of the total variance.

We also compute the cosine similarity between all pairs of vectors connecting the representations of source and target molecules (all $\vv_i, \vv_j$) and we plot a histogram of the similarity distribution (right column in \autoref{fig:pca_plot}). We see that the source-target directions in the representation space are highly aligned. This is also reflected in the average pairwise similarity shown on top of the plot.

Next, we replicate our linear probing strategy, this time using the pairwise data instead of randomly sampled data to train the linear probe. We show the AUC on top of the first column which in both cases is $0.99$. Compared to standard linear probing we get higher performance (see side-by-side comparison in \autoref{sec:appendix_pairwise_probing}). The main reason is that the pairwise probing dataset is balanced by construction making the linear probe easier to train. In contrast, if we randomly sample the probing dataset as before the ratio of molecules with and without a given functional group is highly skewed. Nonetheless, both probing strategies mostly show the same trends when comparing different models and properties.

Under the causal interpretation of pairwise probing we can compute the average treatment effect $\vv_\text{ate}$. We show that it is also highly aligned with the source-target directions as evidenced by the large average cosine similarity (see $C_\text{ate}^\text{pair}$ on top of the middle column in \autoref{fig:pca_plot}). We also study the average treatment effect of removing a functional group on other high-level molecular properties. The results for odor properties are given in \autoref{sec:appendix_causal_smell} due to lack of space. We are able to recover well-known findings about the relations between certain substructures and the smell of a molecule.

\section{RELATED WORK}\label{sec:related_work}
We give a brief overview of probing. For a comprehensive survey, including a discussion of its limitations, see \citet{Belinkov2022ProbingCP}. We also discuss GNN explainability since probing can provide valuable insights on the inductive biases of the models, which can be seen as explanations.

\textbf{Probing.} \label{sec: prob_lit}
The idea of using probing tasks to quantify the information encoded in the representations of a computational system was first proposed by neuroscientists \citep{cox2003functional, kamitani2005decoding, mitchell2008predicting} and was later employed in machine learning research, especially for the study of large language models \citep{AlainB17Understanding, adi2016fine, conneau2018you, liu2019linguistic}.
One line of research is focused on probing via prediction. As we discussed in \autoref{sec:probing_strategies},  selecting the expressive power of the probe is an important choice, with various arguments for linear vs. non-linear probes. \citet{pimentel2020pareto} propose pareto probing to take into account the trade-off between probing accuracy and complexity. \citet{ivanova2021probing} argue that selecting the probe and its complexity depends on the research goal, e.g. do we aim for usability or predictability.
Another line of research studies probing from an information theory perspective. \citet{PimentelVMZWC20Information} suggest that probing measures the mutual information between the property and the representations. \citet{voita2020information} propose a framework based on minimum description length to study how easy is to extract a given property. To measure feature usability \citet{hewitt2021conditional} introduce conditional probing based on $\gV$-information (usable information) \citet{XuZSSE20Usable}. Later \cite{pimentel2021bayesian} extend the notion of usable information and introduce Bayesian probing. 
Finally, \citet{DBLP:conf/naacl/ZhouS21} propose a direct probe that does not rely on classifiers. They consider the geometry of the representation space by using hierarchical clustering.

The overwhelming majority of the probing literature, including the works above, is within the natural language processing domain. There are a few works that study models in computer vision \citep{AlainB17Understanding, ResnickProbing, DBLP:conf/iccv/CaronTMJMBJ21} and biology \citep{Villegas-Morcillo21Unsupervised, RivesMSGLLGOZMF21Biological, ElnaggarProtTrans}. 
The preprint by \citet{Wang22Evaluating} is the only work that considers graph-based models. In contrast to our work, they focus only on self-supervised learning methods, they do not consider transformers-based models, and they rely mostly on probing via prediction.

\textbf{Explainability.} \label{sec: Xgnn_lit}
Most Graph Neural Networks make black box predictions. 
As their use for real-world problems grows, understanding and explaining their predictions become crucial. Explainability methods can be divided into two groups: instance-level and model-level explanations.
There are many instance-level methods based on: gradients, feature attribution, 
perturbations, decomposition, and surrogate models (see \citet{yuan2020explainability} for a comprehensive survey).
%
Model-level explanations are not as well studied. One exception is \citet{yuan2020xgnn} where they employ a generative model to generate the explanation. \citet{zhang2021protgnn} proposes a prototype layer that makes the GNNs self-explainable.
There are also works on counterfactual explanations \citep{lucic2021cf, numeroso2021meg}.
Probing can be seen as a model-level explanation since we quantify how much information about a certain property is encoded in the learned representations.

\section{DISCUSSION AND LIMITATIONS}\label{sec:limitations}
Here, we discuss some of the limitations of the proposed approach and the probing framework in general. For a more detailed discussion see
\citet{Belinkov2022ProbingCP}.
First, the probing results depend on the dataset used for probing and its size. To mitigate this effect we employed incremental probing as suggested by \citet{pimentel2021bayesian}. Moreover, by using the same probing dataset for all models this issue becomes less important  since our main goal is to compare different architectures.
Second, it can be tempting to conclude that because a property $p$ is encoded in the representation $\vz$ it must be important for the prediction of the original model. This is not necessarily the case. For example, \citet{DBLP:conf/eacl/RavichanderBH21} show that models can learn to encode linguistic properties even when they are not needed to solve the main task. To investigate whether the property is important for prediction some authors propose to intervene on $\vz$ to \emph{remove} information about $p$.  \citet{DBLP:journals/tacl/ElazarRJG21} use iterative null space projection to remove information, while \citet{DBLP:journals/coling/FederOSR21} adversarially remove properties. We leave such studies on graph representations for future work.
Third, all of the properties we considered were predefined -- we either compute them without supervision (using RDKit) or use an annotated dataset. Alternatively, as proposed by \citet{DBLP:conf/emnlp/MichaelBT20}, we can look for clusters in the representation space and verify whether they correspond to known properties.

\section{CONCLUSION}
To tackle our main question "What is encoded in the representations learned by graph-based models?" we performed an extensive empirical analysis using the probing framework. We studied five types of properties related to: atoms, meaningful substructures (functional groups), molecular properties (different functions), 3D structure, and odors. We employed several complementary probing strategies, including a pairwise probing strategy that aims at directly isolating the effect of the property. Our findings show that transformer-based models learn richer representations that capture more (chemically) relevant information compared to classical GNNs that use message-passing. Surprisingly, randomly-initialized untrained models also provide useful representations with performance on par with some trained models. We also showed the effect of some design choices, e.g. including skip connections increases the encoded information. We advocate for probing as a debugging tool in model development and as a post-hoc explainability tool.

\bibliography{./main}
\bibliographystyle{plainnat}
\appendix
\input{./supplement.tex}

\end{document}

%% file: math_commands.tex
\usepackage{amsmath,amsfonts,bm}

\def\eqref#1{equation~\ref{#1}}

\def\1{\bm{1}}

\def\vm{{\bm{m}}}

\def\vv{{\bm{v}}}

\def\vx{{\bm{x}}}

\def\vz{{\bm{z}}}

\def\mZ{{\bm{Z}}}

\DeclareMathAlphabet{\mathsfit}{\encodingdefault}{\sfdefault}{m}{sl}
\SetMathAlphabet{\mathsfit}{bold}{\encodingdefault}{\sfdefault}{bx}{n}

\def\gD{{\mathcal{D}}}

\def\gS{{\mathcal{S}}}

\def\gV{{\mathcal{V}}}

\def\sN{{\mathbb{N}}}

\def\sR{{\mathbb{R}}}



%% file: figures/table_linear.tex
\begin{table*}[!t]
\caption{Linear probing performance (AUC score) across models for various functional group properties.}
\label{tab:liinear_probing_models}
\resizebox{\textwidth}{!}{
\begin{tabular}{c|cccccccccc|c}
\hline
                                 & \textbf{Arom. Carb.} & \textbf{Arom. Ring} & \textbf{Satur. Ring} & \textbf{Aniline} & \textbf{Benzene} & \textbf{Bicyclic} & \textbf{Ketone} & \textbf{Methoxy} & \textbf{ParaHydrox.} & \textbf{Pyridine} & \textbf{AVG} \\\hline
\textbf{Raw}   & 83.5 & 97.2 & 84.0 & 78.8 & 83.5 & 89.6 & 81.1 & 80.6 & 83.1 & 82.5 & 84.4 \\    
\textbf{Morgan} & 83.0 & 85.5 & 76.7 & 71.4 & 83.0 & 72.6 & 68.0 & 72.9 & 70.9 & 70.3 & 75.4 \\
\hline
\textbf{GCN}     & 98.1               & 99.3         & 87.8          & 89.9    & 98.1    & 90.1     & 98.1   & 82.3    & 91.1              & 90.4 & 92.5    \\
\textbf{GIN}     & 87.6               & 96.1         & 84.6          & 71.3    & 87.6    & 80.2     & 73.8   & 62.6    & 77.4              & 77.7 & 79.9    \\ \hline
\textbf{Graphormer}      & 99.5               & 99.8         & 90.6          & 98.6    & 99.5    & 95.5     & 99.5   & 86.7    & 96.9              & 98.4 & 96.5    \\
\textbf{GRPE}            & 100.0              & 100.0        & 99.7          & 100.0   & 100.0   & 99.0     & 99.9   & 99.2    & 99.1              & 99.3 & \textbf{99.6}    \\
\textbf{GraphGPS}                 & 99.6               & 100.0        & 92.3          & 98.2    & 99.6    & 96.0     & 99.0   & 93.1    & 97.0              & 95.8 & 97.1    \\
\textbf{TokenGT} & 99.8               & 100.0        & 97.6          & 99.8    & 99.8    & 98.7     & 99.9   & 99.1    & 98.6              & 99.3 & 99.3   \\\hline
\end{tabular}}
\end{table*}

%% file: figures/table_3d_counting.tex
\begin{table*}[!t]
\centering
\caption{Linear probing performance (R2 score) across models for atom and 3D structure properties.}
\label{tab:atom_counting}
\resizebox{\textwidth}{!}{
\begin{tabular}{c|ccccc|ccc}
\hline
\multicolumn{1}{l}{} & \multicolumn{5}{|c}{\textbf{3D Structure Properties}}                           & \multicolumn{3}{|c}{\textbf{Counting Properties}} \\\hline
                     & \textbf{Asphericity} & \textbf{NPR1}    & \textbf{PMI3}     & \textbf{RadiusOfGyration} & \textbf{SpherocityIndex} & \# \textbf{Carbon}   & \# \textbf{Oxygen}  & \# \textbf{Nitrogen}  \\\hline
                     
\textbf{Raw} & 0.152 & 0.131 & 0.734 & 0.827 & 0.174 & 0.033 & -0.850 & 0.000 \\
\textbf{Morgan} & 0.178 & 0.160 & 0.628 & 0.706 & 0.155 &  0.031 & 0.000 & 0.000\\ \hline
                     
\textbf{GCN}          & N/A         & N/A     & N/A     & N/A              & N/A             & 0.586       & N/A        & -6.655       \\
\textbf{GIN}          & -52.922     & -38.858 & -278.344 & -192.431         & -0.016          & 0.555       & -0.037     & -0.016       \\ \hline
\textbf{Graphormer}           & 0.376       & 0.364   & 0.469    & 0.489            & 0.386           & 0.627       & 0.536      & 0.631        \\
\textbf{GRPE}                 & 0.490       & 0.457   & 0.757    & 0.874            & 0.469           & 0.729       & 0.196      & 0.221        \\
\textbf{GraphGPS}             & 0.268       & 0.275   & 0.803    & 0.845            & 0.332           & 0.898       & 0.746      & 0.806        \\
\textbf{TokenGT}      & 0.487       & 0.490   & 0.757    & 0.779            & 0.433           & 0.710       & 0.665      & 0.769       \\\hline
\end{tabular}
}
\end{table*}

%% file: supplement.tex
\onecolumn
\aistatstitle{Appendix}
\section{ADDITIONAL EXPERIMENTS AND DETAILS}
\label{sec:appendix_additional_experiments}
The appendix is structured as follows. First, we provide details on how our baselines are trained. In \autoref{sec:appendix_train_eval_linear_probes}, we provide additional details on how we train our linear probes for different probing tasks. In \autoref{sec:app_smells} we provide study odor probing in more detail.
In \autoref{sec:app_additional_probing_perf}, we extend our probing framework to the additional tasks and provide further insights. In \autoref{sec:appendix_molecule_stats}, we provide the statistics for our pretraining datasets. In \autoref{sec:app_over_smoothing}, we analyze the effect of the over-smoothing phenomenon for different baselines and present our observations. \autoref{sec:appendix_pairwise_probing}, \autoref{sec:appendix_causal_smell}, and \autoref{appendix: overlap} reflect detailed studies on pairwise probing, the causal effect of functional groups, and overlap between pre-training datasets respectively. 


\subsection{Training and Evaluating Linear Probes}
\label{sec:appendix_train_eval_linear_probes}
We train a logistic regression model for categorical properties where $p_i \in \{0, \dots, C\}$ where $C$ is the number of categories, e.g., high-level molecular prediction, and meaningful substructures. We standardize the representation of each model by removing the mean and scaling it to unit variance to increase the convergence rate of the logistic regression.  
For tasks that contain positive integers as labels ($p_i \in \sN, p_i \geq 0$.), e.g. counting atom properties, we train linear Poisson regression probes. Since the 3D structure properties are continuous we train a standard linear regression probe.
%
%
For pooling, all our baseline models use virtual nodes unless mentioned otherwise.

\subsection{Odor Probing}
\label{sec:app_smells}

The objective of this experiment is to probe for the presence of informative signals about the smell (odor) of a molecule, e.g. sweet or woody (32 smells in total) in the learned representations, using the Pyrfume data archive \footnote{Each dataset and the corresponding reference can be found on \url{https://github.com/pyrfume/pyrfume-data}.}. As before, we pretrain models on the PCQM4Mv2 dataset and we apply linear probing. We split the dataset into 80\% training and 20\% testing using the stratified split for each smell.
We tabulate and present the results in \autoref{tab:app_smell}.
Again, we see that on average the representations learned by transformer-based models encode more information about odors. GIN is outperformed by GCN even though it is theoretically more expressive.
Certain odors (e.g. sulfurous) are reliably encoded by all models.

\begin{table}[h]
\centering
\caption{Linear probing performance (AUC score) for different odors.}
\label{tab:app_smell}
\resizebox{\textwidth}{!}{
\begin{tabular}{c|ccccccccccccccccc}
\textbf{Model}      & \textbf{Apple} & \textbf{Balsamic} & \textbf{Burnt} & \textbf{Caramellic} & \textbf{Cheesy} & \textbf{Citrus} & \textbf{Earthy} & \textbf{Ethereal} & \textbf{Fatty} & \textbf{Fermented} & \textbf{Floral} & \textbf{Fresh} & \textbf{Fruity} & \textbf{Green} & \textbf{Herbal} & \textbf{Meaty} & \textbf{Mint} \\\hline
\textbf{Raw} & 0.52 & 0.57 & 0.55 & 0.55 & 0.60 & 0.60 & 0.53 & 0.65 & 0.57 & 0.54 & 0.61 & 0.50 & 0.74 & 0.67 & 0.57 & 0.65 & 0.61 \\
\textbf{Morgan} & 0.55 & 0.64 & 0.50 & 0.52 & 0.52 & 0.60 & 0.51 & 0.59 & 0.66 & 0.55 & 0.61 & 0.53 & 0.76 & 0.70 & 0.55 & 0.55 & 0.60 \\ \hline
\textbf{GCN}        & 0.91           & 0.88              & 0.90           & 0.93                & 0.92            & 0.85            & 0.78            & 0.91              & 0.88           & 0.91               & 0.83            & 0.74           & 0.82            & 0.79           & 0.72            & 0.89           & 0.89          \\
\textbf{GIN}        & 0.80           & 0.88              & 0.85           & 0.77                & 0.79            & 0.83            & 0.63            & 0.90              & 0.74           & 0.87               & 0.76            & 0.65           & 0.71            & 0.71           & 0.65            & 0.88           & 0.75          \\
\textbf{Graphormer} & 0.88           & 0.87              & 0.85           & 0.84                & 0.90            & 0.81            & 0.67            & 0.93              & 0.87           & 0.93               & 0.76            & 0.70           & 0.81            & 0.79           & 0.71            & 0.92           & 0.84          \\
\textbf{GRPE}       & 0.88           & 0.91              & 0.91           & 0.88                & 0.88            & 0.87            & 0.69            & 0.91              & 0.86           & 0.92               & 0.83            & 0.72           & 0.83            & 0.80           & 0.73            & 0.94           & 0.87          \\
\textbf{GraphGPS}   & 0.89           & 0.87              & 0.86           & 0.92                & 0.89            & 0.82            & 0.75            & 0.93              & 0.90           & 0.89               & 0.81            & 0.74           & 0.82            & 0.79           & 0.76            & 0.92           & 0.91          \\
\textbf{TokenGT}    & 0.90           & 0.94              & 0.90           & 0.88                & 0.87            & 0.84            & 0.73            & 0.93              & 0.89           & 0.90               & 0.79            & 0.74           & 0.81            & 0.79           & 0.78            & 0.90           & 0.90  \\\hline     
\end{tabular}
}
\\
\resizebox{\textwidth}{!}{
\begin{tabular}{c|ccccccccccccccc|c}
     & \textbf{Nutty} & \textbf{Oily} & \textbf{Onion} & \textbf{Pineapple} & \textbf{Pungent} & \textbf{Roasted} & \textbf{Rose} & \textbf{Spicy} & \textbf{Sulfurous} & \textbf{Sweet} & \textbf{Tropical} & \textbf{Vegetable} & \textbf{Waxy} & \textbf{Winey} & \textbf{Woody} & \textbf{AVG} \\\hline
\textbf{Raw} & 0.66 & 0.62 & 0.71 & 0.53 & 0.52 & 0.62 & 0.57 & 0.57 & 0.71 & 0.56 & 0.61 & 0.55 & 0.56 & 0.51 & 0.58 & 0.59 \\
\textbf{Morgan} & 0.55 & 0.66 & 0.66 & 0.57 & 0.57 & 0.52 & 0.65 & 0.57 & 0.57 & 0.57 & 0.52 & 0.51 & 0.65 & 0.57 & 0.50 & 0.57 \\ \hline

\textbf{GCN}        & 0.78           & 0.88          & 0.90           & 0.93               & 0.83             & 0.89             & 0.88          & 0.80           & 0.95               & 0.69           & 0.81              & 0.81               & 0.92          & 0.84           & 0.81           & 0.85         \\
\textbf{GIN}        & 0.78           & 0.91          & 0.92           & 0.84               & 0.89             & 0.89             & 0.77          & 0.78           & 0.90               & 0.63           & 0.65              & 0.73               & 0.82          & 0.79           & 0.74           & 0.79         \\
\textbf{Graphormer} & 0.76           & 0.80          & 0.91           & 0.90               & 0.85             & 0.90             & 0.90          & 0.81           & 0.96               & 0.64           & 0.84              & 0.84               & 0.83          & 0.81           & 0.80           & 0.83         \\
\textbf{GRPE}       & 0.82           & 0.83          & 0.94           & 0.90               & 0.87             & 0.91             & 0.92          & 0.79           & 0.96               & 0.67           & 0.86              & 0.84               & 0.88          & 0.82           & 0.81           & 0.85         \\
\textbf{GraphGPS}   & 0.81           & 0.91          & 0.94           & 0.94               & 0.91             & 0.93             & 0.84          & 0.83           & 0.97               & 0.69           & 0.79              & 0.84               & 0.93          & 0.81           & 0.81           & \textbf{0.86}         \\
\textbf{TokenGT}    & 0.79           & 0.88          & 0.96           & 0.93               & 0.85             & 0.92             & 0.89          & 0.78           & 0.98               & 0.70           & 0.83              & 0.83               & 0.87          & 0.86           & 0.81           & 0.85        
\end{tabular}
}

\end{table}

\subsection{Additional Probing Performance}
\label{sec:app_additional_probing_perf}
\textbf{High-Level Molecular Properties.} In the main paper (\autoref{tab:transfer}) we presented a subset of probing results for the high-level molecular properties. In \autoref{tab:transfer_app} we show for completeness the detailed analysis for all high-level molecular properties from the MoleculeNet benchmark. 

\begin{table}[!h]
\caption{Linear probing performance (AUC and R2 scores) for high-level molecular properties from MoleculeNet.}
\label{tab:transfer_app}
\centering
\resizebox{0.7\textwidth}{!}{
\begin{tabular}{ccccccccc|c}
\hline
\textbf{Dataset}         & \textbf{BBBP}  & \textbf{Tox21} & \textbf{ToxCast} & \textbf{Sider} & \textbf{ClinTox} & \textbf{MUV}    & \textbf{HIV}    & \textbf{BACE}   & \textbf{AVG}   \\ \hline
\textbf{Molecules}       & 2,039 & 7,831 & 8,575   & 1,427 & 1,478   & 93,087 & 41,127 & 1,5113 &       \\
\textbf{Tasks}           & 1     & 12    & 617     & 27    & 2       & 17     & 1      & 1      &       \\ \hline
\textbf{Raw}  & 66.8 & 68.1 & 60.5 & 56.2 & 60.6 & 69.5 & 73.2 & 75.1 & 66.3  \\
\textbf{Morgan}  & 63.2 & 64.0 & 54.1 & 60.5 & 57.4 & 67.2 & 65.8 & 75.0 & 63.4 \\ \hline
\textbf{GCN}     & 56.5  & 60.2  & 57.0    & 55.2  & 64.9    & 72.9   & 66.3   & 71.1   & 63.0  \\
\textbf{GIN}     & 58.8  & 58.1  & 55.7    & 53.2  & 57.0    & 64.6   & 72.4   & 77.1   & 62.1  \\
\textbf{Graphormer}      & 59.6  & 69.0  & 59.2    & 57.5  & 78.5    & 75.7   & 75.8   & 63.0   & 67.2  \\
\textbf{GRPE}            & 64.9  & 72.1  & 61.1    & 55.7  & 82.8    & 75.7   & 73.0   & 77.9   & 70.4  \\
\textbf{GraphGPS}        & 63.9  & 70.3  & 60.0    & 60.7  & 72.7    & 74.4   & 72.2   & 74.5   & 68.5  \\
\textbf{TokenGT} & 57.1	& 70.8	& 59.3	  & 58.4  &	88.5 	& 80.4	 & 75.8	  & 75.2   & \textbf{70.6}  \\ \hline
\end{tabular}
}
\end{table}

\textbf{Impact of the Pretraining Dataset.}
Similarly, in \autoref{tab:dataset_prob_mol} we presented a subset of the results for the impact of the pretraining dataset and here we show all properties.
Since the number of training samples in the \textit{ogbg-molpcba-subset} and \textit{ogbg-molhiv} datasets are the same, their performance can be interpreted as the impact of multitask learning vs. single-task pretraining. We bank on the assumption that equalizing the training sample size would help us study the effect of the number of tasks. \autoref{tab:dataset_prob_prop} suggest that probing performance increases significantly when we use the \textit{ogbg-molpcba-subset} dataset instead of \textit{ogbg-molhiv} dataset. This implies that pretraining on multitask learning increases the information related to topological properties of the graph, which in this case is the existence of functional groups. 

\begin{table*}[h]
\caption{Effect of the pretraining dataset on probing performance (AUC score).}
\label{tab:dataset_prob_prop}
\resizebox{\textwidth}{!}{
\begin{tabular}{c|c|cccccccccc|c}
\hline
\textbf{Architecture}                          & \textbf{Dataset}                      & \textbf{Arom. Carb.} & \textbf{Arom. Ring.} & \textbf{Satur. Ring.} & \textbf{Aniline} & \textbf{Benzene} & \textbf{Bicyclic} & \textbf{Ketone} & \textbf{Methoxy} & \textbf{ParaHydrox} & \textbf{Pyridine} & \textbf{AVG} \\ \hline
\multirow{3}{*}{\textbf{GCN}}         & \textbf{Molhiv}         & 98.6               & 99.5         & 95.3          & 83.8    & 98.5    & 85.6     & 87.2   & 97.7    & 89.1              & 87.9     &  \textbf{93.3}    \\
                                      & \textbf{Molpcba-sub} & 99.7               & 99.8         & 98.6          & 95.3    & 99.7    & 94.2     & 97.2   & 99.6    & 94.2              & 96.0     &  \textbf{97.4}   \\
                                      & \textbf{Molpcba}        & 99.6               & 99.7         & 98.1          & 96.3    & 99.6    & 95.5     & 96.8   & 99.0    & 94.4              & 95.6     &   \textbf{97.5}  \\\hline
\multirow{3}{*}{\textbf{GIN}}         & \textbf{Molhiv}         & 82.0               & 90.3         & 83.3          & 65.2    & 81.9    & 66.5     & 63.5   & 71.1    & 73.5              & 71.2     &  74.9   \\
                                      & \textbf{Molpcba-sub} & 97.6               & 99.7         & 97.5          & 91.1    & 97.6    & 94.7     & 91.4   & 95.2    & 90.5              & 91.4     &  94.7   \\
                                      & \textbf{Molpcba}        & 99.2               & 99.8         & 98.1          & 95.6    & 99.2    & 95.5     & 94.3   & 95.5    & 91.5              & 94.5     &  96.3   \\\hline
\multirow{3}{*}{\textbf{V-GCN}} & \textbf{Molhiv}         & 94.5               & 97.3         & 90.8          & 76.2    & 94.5    & 73.7     & 76.8   & 88.4    & 78.1              & 76.7     &  84.7   \\
                                      & \textbf{Molpcba-sub} & 98.3               & 99.5         & 95.1          & 90.2    & 98.3    & 87.2     & 91.5   & 93.0    & 83.3              & 92.1     &  92.6   \\
                                      & \textbf{Molpcba}        & 98.7               & 99.5         & 95.4          & 93.3    & 98.7    & 89.4     & 95.0   & 92.9    & 87.6              & 93.0     &  94.4   \\\hline
\multirow{3}{*}{\textbf{V-GIN}} & \textbf{Molhiv}         & 85.4               & 92.5         & 87.4          & 71.3    & 85.4    & 68.5     & 69.5   & 70.5    & 74.7              & 73.0     &  77.8   \\
                                      & \textbf{Molpcba-sub} & 94.5               & 99.7         & 95.1          & 81.8    & 94.6    & 88.3     & 84.4   & 84.7    & 81.6              & 85.8     &  89.0  \\
                                      & \textbf{Molpcba}        & 98.3               & 99.8         & 96.2          & 90.5    & 98.3    & 91.2     & 90.9   & 90.3    & 86.9              & 92.1     &  93.5  \\\hline
\end{tabular} 
}
\end{table*}


\textbf{Impact of Design Choices.} 
To complement the results we show in \autoref{fig: abl-DC} on the impact of various architectural design choices we report additional results in \autoref{tab:arch_design_choice}. As before, adding residual connections (models with ``Res'') improves the probing performance. For GIN using a virtual node improves performance, while for GCN it has the opposite effect. Using jumping knowledge (models with ``JK'') also tends to improve the performance.

\begin{table}[h]
\caption{Impact of using residual connections (models with ``Res''), pooling with or without virtual nodes, and jumping knowledge (models with ``JK'') on the linear probing performance (AUC score).}
\label{tab:arch_design_choice}
\centering
\resizebox{\textwidth}{!}{
\begin{tabular}{ccccccccccc|c}
\hline
                            & \textbf{AromaticCarbocycle} & \textbf{AromaticRing} & \textbf{SaturatedRing} & \textbf{Aniline} & \textbf{Benzene} & \textbf{Bicyclic} & \textbf{Ketone} & \textbf{Methoxy} & \textbf{ParaHydroxylation} & \textbf{Pyridine} & \textbf{AVG} \\\hline
\textbf{GCN}                & 0.98               & 0.99         & 0.94          & 0.79    & 0.98    & 0.83     & 0.82   & 0.94    & 0.87              & 0.87  & 0.90    \\
\textbf{GCN-JK}             & 0.98               & 0.99         & 0.98          & 0.88    & 0.98    & 0.88     & 0.92   & 0.97    & 0.88              & 0.91    & 0.94     \\
\textbf{GCN-Res}            & 0.98               & 0.99         & 0.96          & 0.86    & 0.98    & 0.87     & 0.91   & 0.97    & 0.88              & 0.89  & 0.93   \\
\textbf{GCN-Res-JK}         & 0.99               & 0.99         & 0.97          & 0.90    & 0.99    & 0.89     & 0.93   & 0.98    & 0.90              & 0.90  & 0.94   \\
\textbf{Virtual GCN}        & 0.94               & 0.97         & 0.91          & 0.73    & 0.94    & 0.73     & 0.74   & 0.83    & 0.77              & 0.76    & 0.83     \\
\textbf{Virtual GCN-JK}     & 0.98               & 0.99         & 0.97          & 0.89    & 0.98    & 0.85     & 0.88   & 0.96    & 0.85              & 0.87     & 0.92  \\
\textbf{Virtual GCN-Res}    & 0.98               & 0.99         & 0.96          & 0.88    & 0.98    & 0.84     & 0.86   & 0.93    & 0.86              & 0.84   & 0.91  \\
\textbf{Virtual GCN-Res-JK} & 0.98               & 0.99         & 0.97          & 0.90    & 0.98    & 0.85     & 0.91   & 0.95    & 0.87              & 0.87    & 0.93  \\\hline
\textbf{GIN}                & 0.79               & 0.86         & 0.82          & 0.63    & 0.79    & 0.64     & 0.62   & 0.68    & 0.71              & 0.68    & 0.72  \\
\textbf{GIN-JK}             & 0.98               & 0.99         & 0.97          & 0.90    & 0.98    & 0.90     & 0.91   & 0.97    & 0.88              & 0.89     & 0.94 \\
\textbf{GIN-Res}            & 0.99               & 1.00         & 0.97          & 0.92    & 0.99    & 0.90     & 0.92   & 0.98    & 0.91              & 0.90     & \textbf{0.95}\\
\textbf{GIN-Res-JK}                  & 0.99               & 1.00         & 0.97          & 0.91    & 0.99    & 0.90     & 0.91   & 0.97    & 0.90              & 0.89    &0.94 \\
\textbf{Virtual GIN}        & 0.84               & 0.91         & 0.87          & 0.70    & 0.84    & 0.66     & 0.67   & 0.66    & 0.72              & 0.72    &0.76 \\
\textbf{Virtual GIN-JK}     & 0.98               & 0.99         & 0.96          & 0.87    & 0.98    & 0.84     & 0.85   & 0.91    & 0.85              & 0.83    &0.91 \\
\textbf{Virtual GIN-Res}    & 0.98               & 0.99         & 0.96          & 0.89    & 0.98    & 0.86     & 0.87   & 0.93    & 0.87              & 0.85     &0.92 \\
\textbf{Virtual GIN-Res-JK} & 0.98               & 1.00         & 0.96          & 0.89    & 0.98    & 0.86     & 0.86   & 0.92    & 0.86              & 0.84    &0.92   \\\hline
\end{tabular}
}
\end{table}

\newpage
\subsection{Datasets Molecules Statistics}
\label{sec:appendix_molecule_stats}
We use ogbg-molhiv and ogbg-molpcba to compare the effect of the pretraining dataset on the performance of the probing task since both contain small molecules with similar statistics. \autoref{tab: statistic} shows the average number of nodes and edges per graph which indicate that both datasets contain small molecules with similar sparsity.   

\begin{table}[!h]
\caption{The molecules statistic for Molhiv and Molpcba dataset}
\label{tab: statistic}
\centering
\begin{tabular}{c|ccc}
\textbf{Name}         & \textbf{\#Graphs} & \textbf{\#Nodes per graph} & \textbf{\#Edges per graph} \\\hline
\textbf{ogbg-molhiv}  & 41,127   & 25.5              & 27.5              \\
\textbf{ogbg-molpcba} & 437,929  & 26.0              & 28.1             
\end{tabular}
\end{table}

\subsection{Oversmoothing Analysis}
\label{sec:app_over_smoothing}
\begin{figure}[!h]
\centering 
\includegraphics[width=0.4\linewidth]{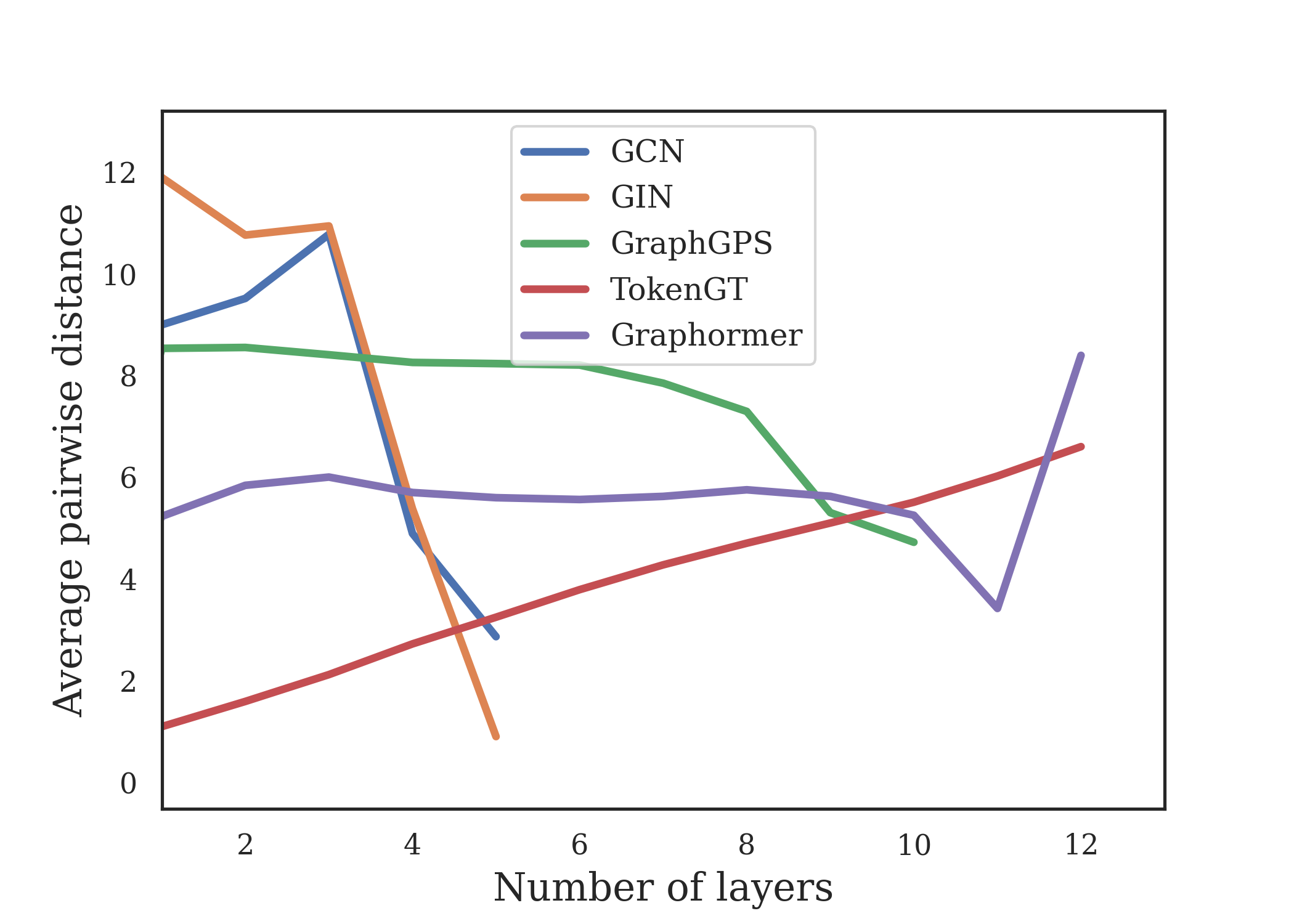}
\caption{Effect of oversmoothing across several models.}
\label{fig:oversmoothing}
\end{figure}

We compare the learned feature embedding of various models to see whether the oversmoothing phenomenon is responsible for degradation in probing performance. In \autoref{fig:oversmoothing}, we compute the average pairwise distance between the (learned) node features and average it across all the molecules. For models like GCN-virtual and GIN-virtual, we observe that as the number of layers (hops) increases, the average distance shrinks. The transformer-based models are not as susceptible to this phenomenon. In any case, we cannot conclude that the decrease in probing performance is due to oversmoothing.    

\subsection{Overlap Between MoleculeNet and MOLPCBA datasets}
\label{appendix: overlap}

\autoref{tab: overlap} show the overlap between the Molecule Net datasets and ogbg-molpcba dataset. Some datasets (e.g. BACE) have no or minimial overlap, while others (e.g. MUV) have a substantial overlap.
\begin{table}[t]
\caption{The overlap between Molpcba molecule and MoleculeNet datasets}
\label{tab: overlap}
\centering
\begin{tabular}{c|cc}
      \textbf{Dataset}           & \textbf{Number of sample} & \textbf{Overlap with MOLPCBA} \\ \hline
\textbf{BACE}    & 1513              & 0                             \\ 
\textbf{BBBP}    & 2039              & 88                            \\ 
\textbf{ClinTox} & 1477              & 22                            \\ 
\textbf{MUV}     & 93087             & 89748                         \\ 
\textbf{Sider}   & 1427              & 14                            \\ 
\textbf{Tox21}   & 7831              & 4687                          \\ 
\textbf{ToxCast} & 8576              & 570                           \\ 
\textbf{HIV}  & 41127             & 1943                          \\ 
\end{tabular}
\end{table}

\subsection{Pairwise Probing}
\label{sec:appendix_pairwise_probing}
\begin{figure}[!h]
    \centering
    
    \includegraphics[width=0.85\textwidth]{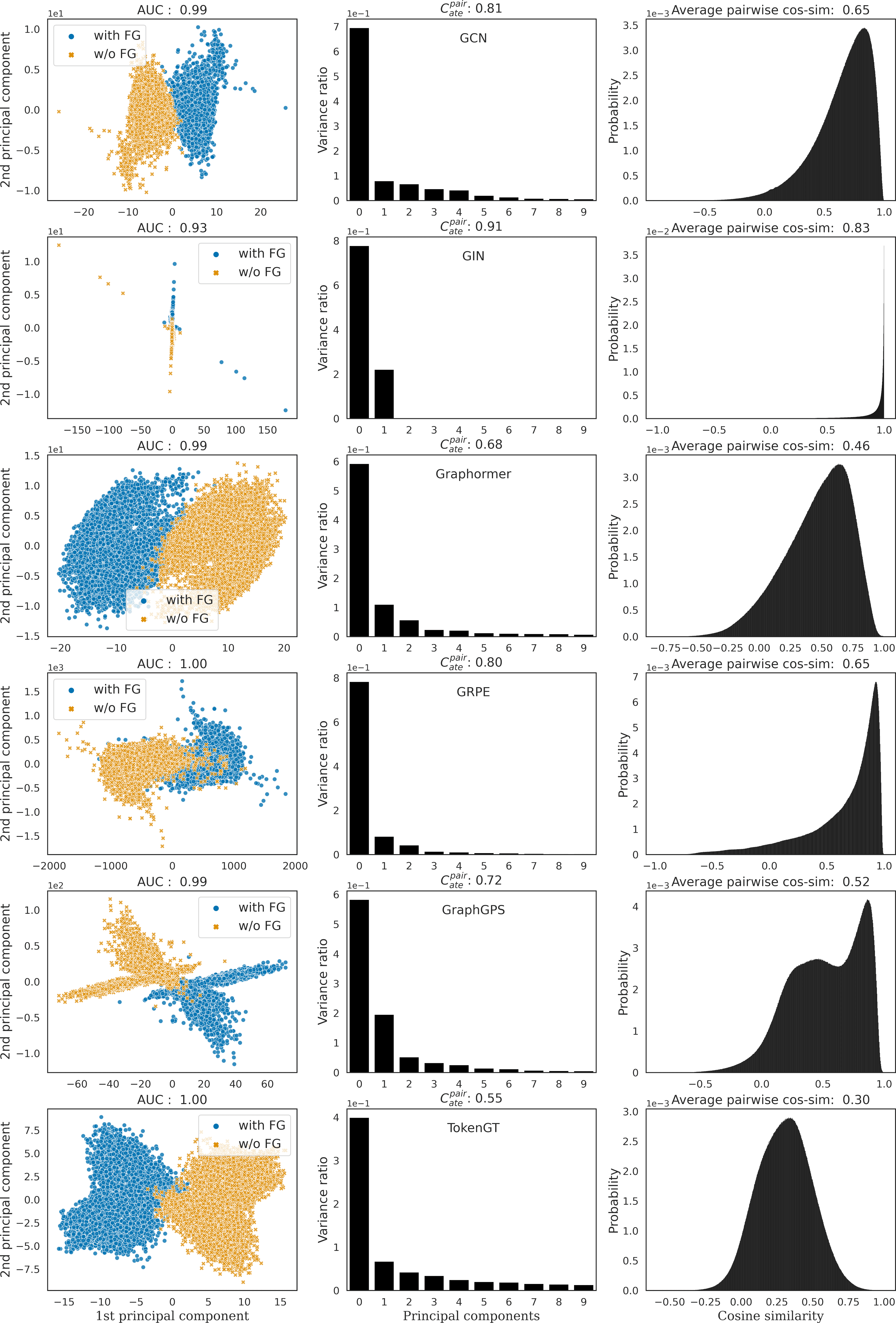}
    \caption{Pairwise probing for various architecture for Nitro functional group. We show the projection on the top two principal components (left), the ratio of explained variance per component (middle), and the pairwise cosine similarity (right).}
    \label{fig: appendix_Nitro_all}
\end{figure}

\begin{figure}[!h]
    \centering
    
    \includegraphics[width=0.85\textwidth]{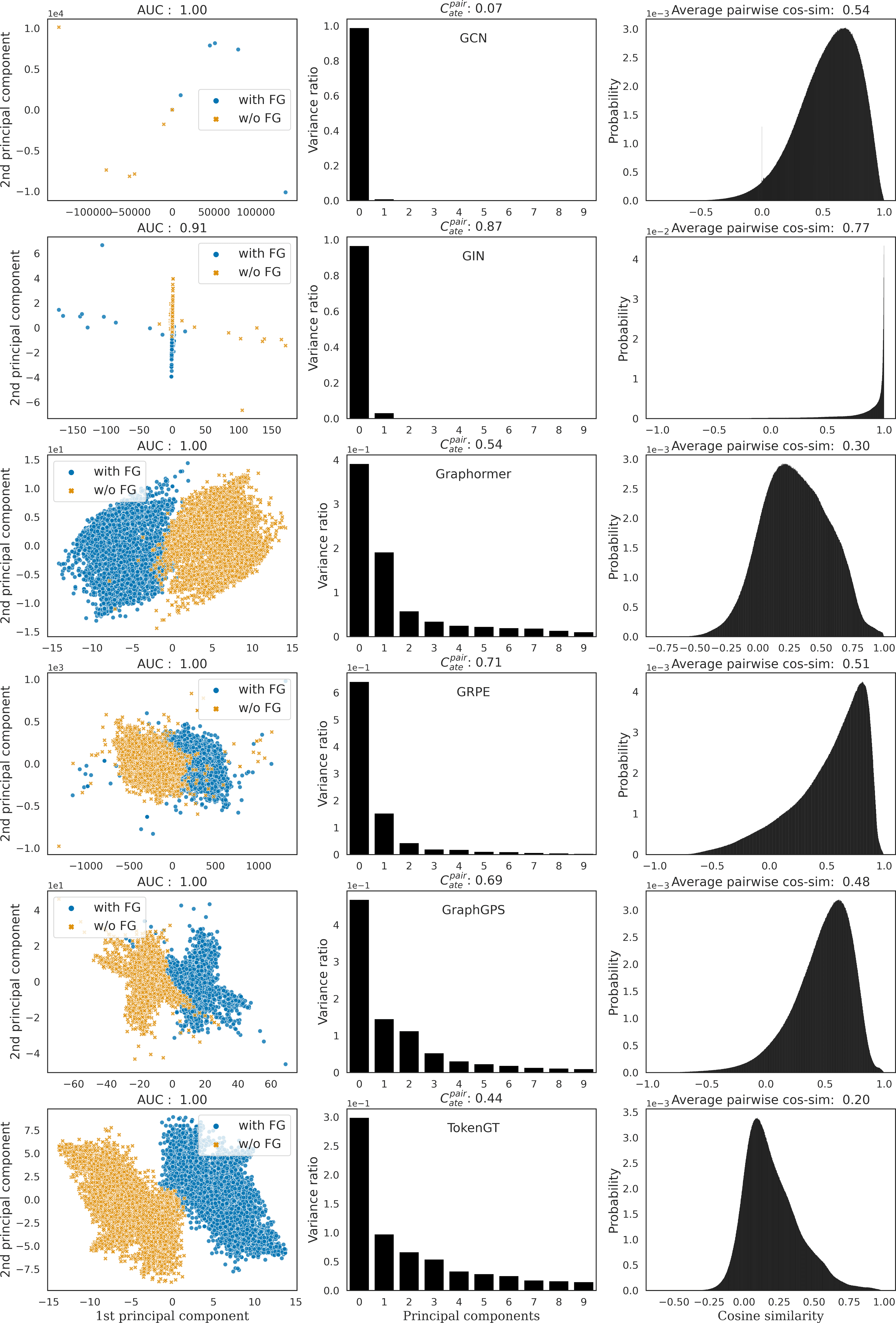}
    \caption{Pairwise probing for various architecture for Amide functional group. We show the projection on the top two principal components (left), the ratio of explained variance per component (middle), and the pairwise cosine similarity (right).}
    \label{fig: appendix_amide_all}
\end{figure}

We provide additional results for pairwise probing on all the models for Nitro (\autoref{fig: appendix_Nitro_all}) and Amide (\autoref{fig: appendix_amide_all}) functional groups. The results and observations are similar as in the main paper. The top two principal components are enough to reliably identify the property of interest for most models. This is also reflected in the ratio of the captured variance.  

\subsection{Causal Effect on Odor}
\label{sec:appendix_causal_smell}

We also extend our pairwise probing for studying the causal effect of the functional group on high-level properties. In this case, we study the effect of a functional group ($F$) on the smell of a molecule ($Y$). In this regard, we select a source molecule that has a specific smell and a given functional group, and then we create the target molecule in the pair by removing the functional group. Next, we train a probe on the representations of the pre-trained model to predict specific smells. Then, we use our probe models and molecule pairs (with and without specific functional groups) to measure the effect of a functional group on predicting specific smells. Based on the SCM we assumed in the main paper, since removing the functional group is an intervention and removes the edge from E to F, the change in the prediction probability is the causal effect of the functional group on the smell. 

\begin{figure}[!h]
    
    \centering
    \includegraphics[width=0.98\textwidth]{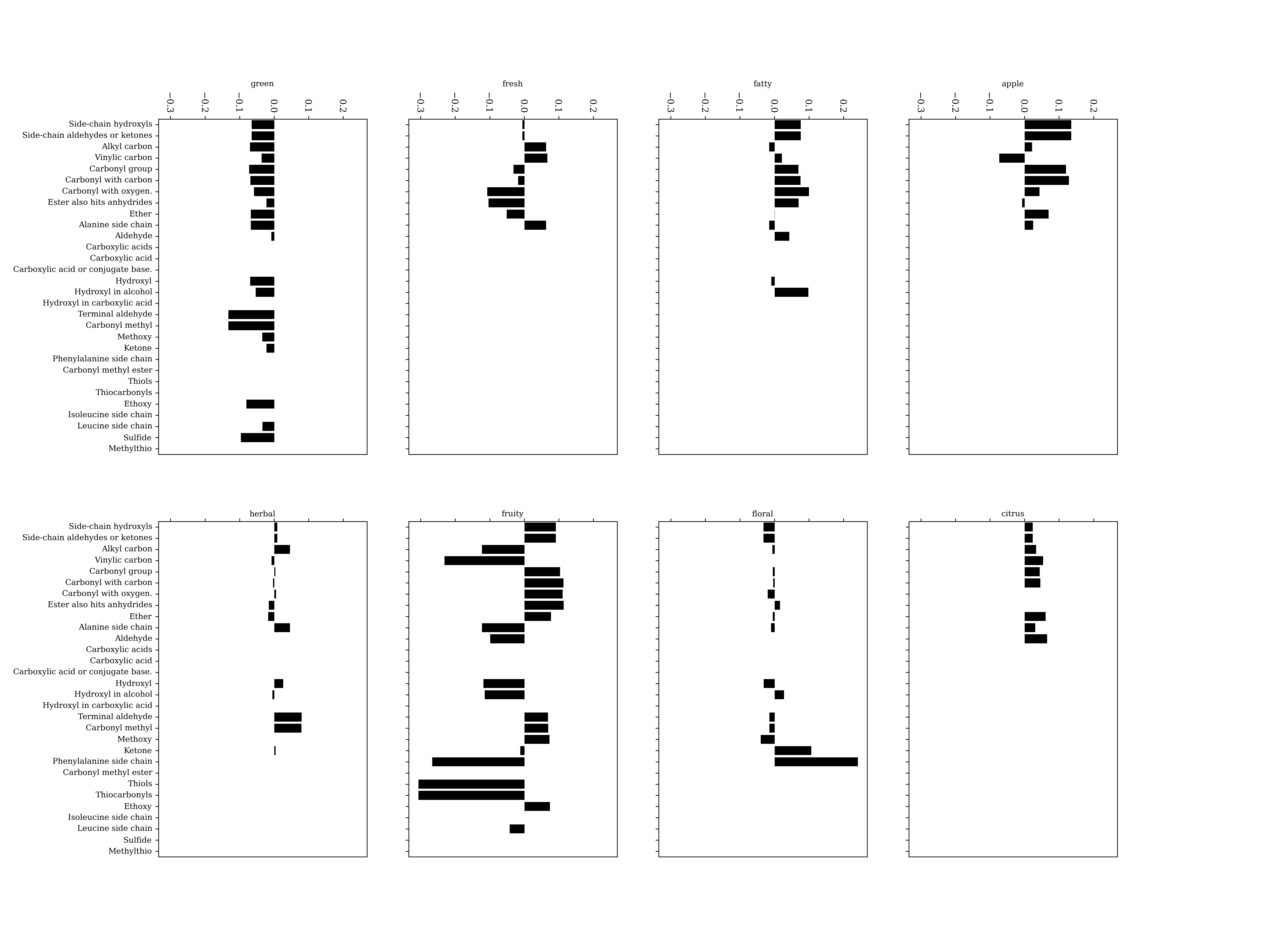}

    \caption{Causal effect of removing a functional group on the predicted odor of a molecule.}
    \label{fig: causal_effect_fg_smell}
\end{figure}

\begin{figure}[!h]
    
    \centering
    
    \includegraphics[width=0.98\textwidth]{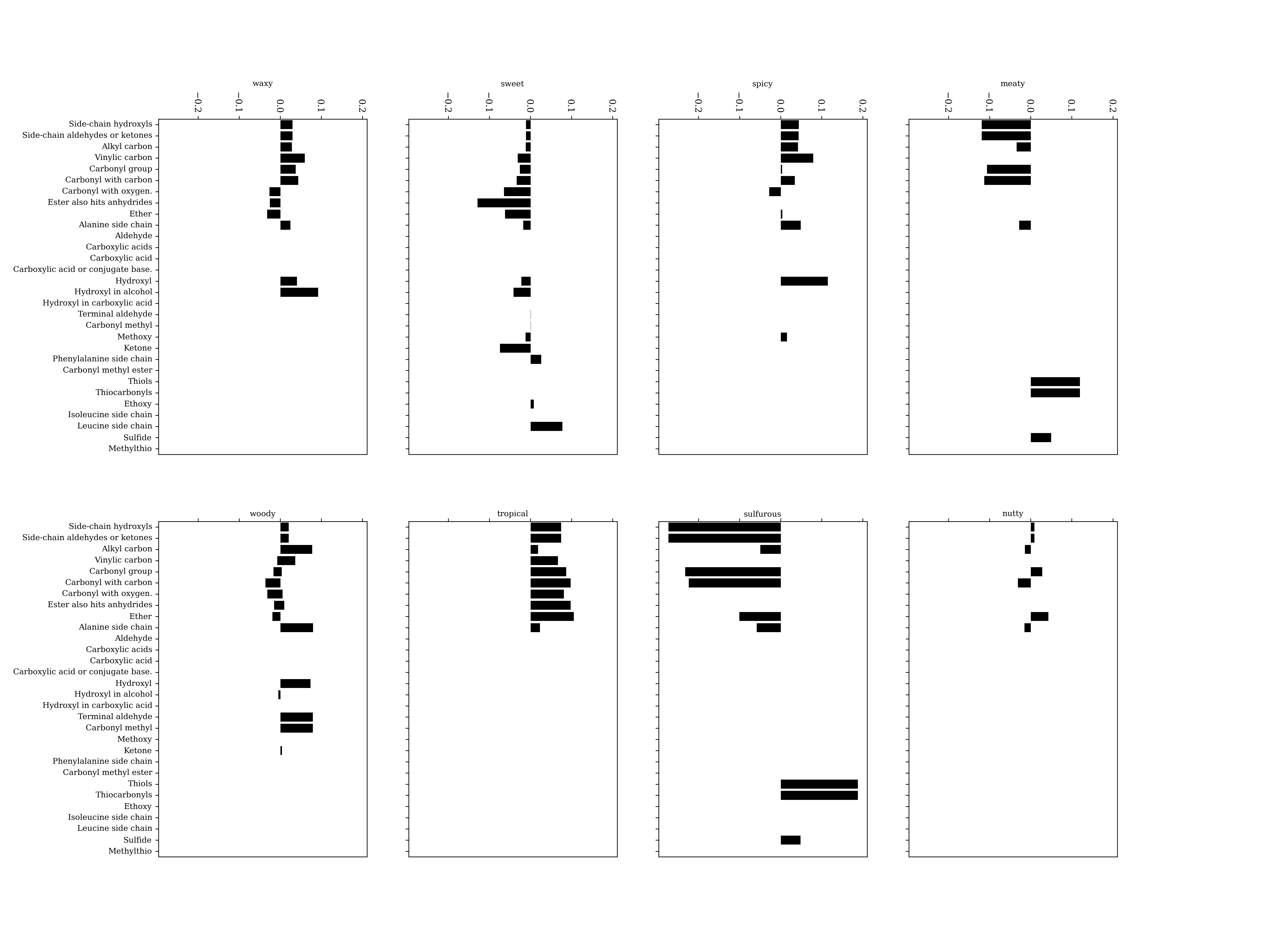}
    
    \caption{Causal effect of removing a functional group on the predicted odor of a molecule.}
    \label{fig: causal_effect_fg_smell2}
\end{figure}

\autoref{fig: causal_effect_fg_smell}, \autoref{fig: causal_effect_fg_smell2} show results for 16 different odors and 30 different functional groups. Here, we subtracted the mean across functional groups to account for the general effect of removing any substructure from a molecule.
Although the result is noisy, there are several chemically meaningful signals. For example, the compounds that contain sulfide are sweet, or there is a relation between the ketone and the apple smell, which seems to be interesting for further study.

